\documentclass{article} %
\usepackage{iclr2027_conference,times}

\usepackage{amsmath,amsfonts,bm}

\def\eqref#1{equation~\ref{#1}}

\def\1{\bm{1}}

\DeclareMathAlphabet{\mathsfit}{\encodingdefault}{\sfdefault}{m}{sl}
\SetMathAlphabet{\mathsfit}{bold}{\encodingdefault}{\sfdefault}{bx}{n}

\usepackage{amsmath}
\usepackage{amssymb}
\usepackage{booktabs}
\usepackage{multirow}
\usepackage{tabularx}
\usepackage{graphicx}
\usepackage{enumitem}
\usepackage{float}
\usepackage{capt-of}
\usepackage{wrapfig}
\usepackage{needspace}
\usepackage{xcolor}
\usepackage{titletoc}
\usepackage{hyperref}
\hypersetup{hidelinks}
\usepackage{url}
\usepackage{listings}
\usepackage{tcolorbox}
\tcbuselibrary{listings,breakable,skins}

\DeclareMathSizes{7.5}{7}{5}{5}

\title{Learning to Prove, Not Just to Answer:\\
Reinforcement Learning from Formal Verification
for Natural-Language Logical Reasoning\par\vspace{0.12in}}

\author{\normalfont\normalsize
\renewcommand{\arraystretch}{1.20}%
\begin{tabular}{@{}l@{}}
\textbf{Qili Zhang}\textsuperscript{1,2}\quad
\textbf{Qianren Mao}\textsuperscript{1}\quad
\textbf{Hanze Cai}\textsuperscript{2}\quad
\textbf{Kaiming Zhao}\textsuperscript{2}\quad
\textbf{Yuening He}\textsuperscript{2} \\
\textbf{Xihan Lei}\textsuperscript{2}\quad
\textbf{Yashuo Luo}\textsuperscript{2}\quad
\textbf{Hanwen Hao}\textsuperscript{2}\quad
\textbf{Yutong Gu}\textsuperscript{3}\quad
\textbf{Likang Xiao}\textsuperscript{2} \\
\textbf{Zhijun Chen}\textsuperscript{4}\quad
\textbf{Weifeng Jiang}\textsuperscript{3}\quad
\textbf{Haoyi Zhou}\textsuperscript{2}\quad
\textbf{Jianxin Li}\textsuperscript{2} \\
\textsuperscript{1}Zhongguancun Laboratory\qquad
\textsuperscript{2}Beihang University \\
\textsuperscript{3}Nanyang Technological University\qquad
\textsuperscript{4}Hong Kong Polytechnic University
\end{tabular}}

\iclrfinalcopy

\newcommand{\method}{\textsc{Proof-R1}}
\newcommand{\MCFV}{MCFV}
\newcommand{\ASDC}{ASDC}
\newlength{\captioniconheight}
\DeclareRobustCommand{\captionicon}{%
  \begingroup
  \settoheight{\captioniconheight}{X}%
  \raisebox{0pt}{\includegraphics[height=\captioniconheight]{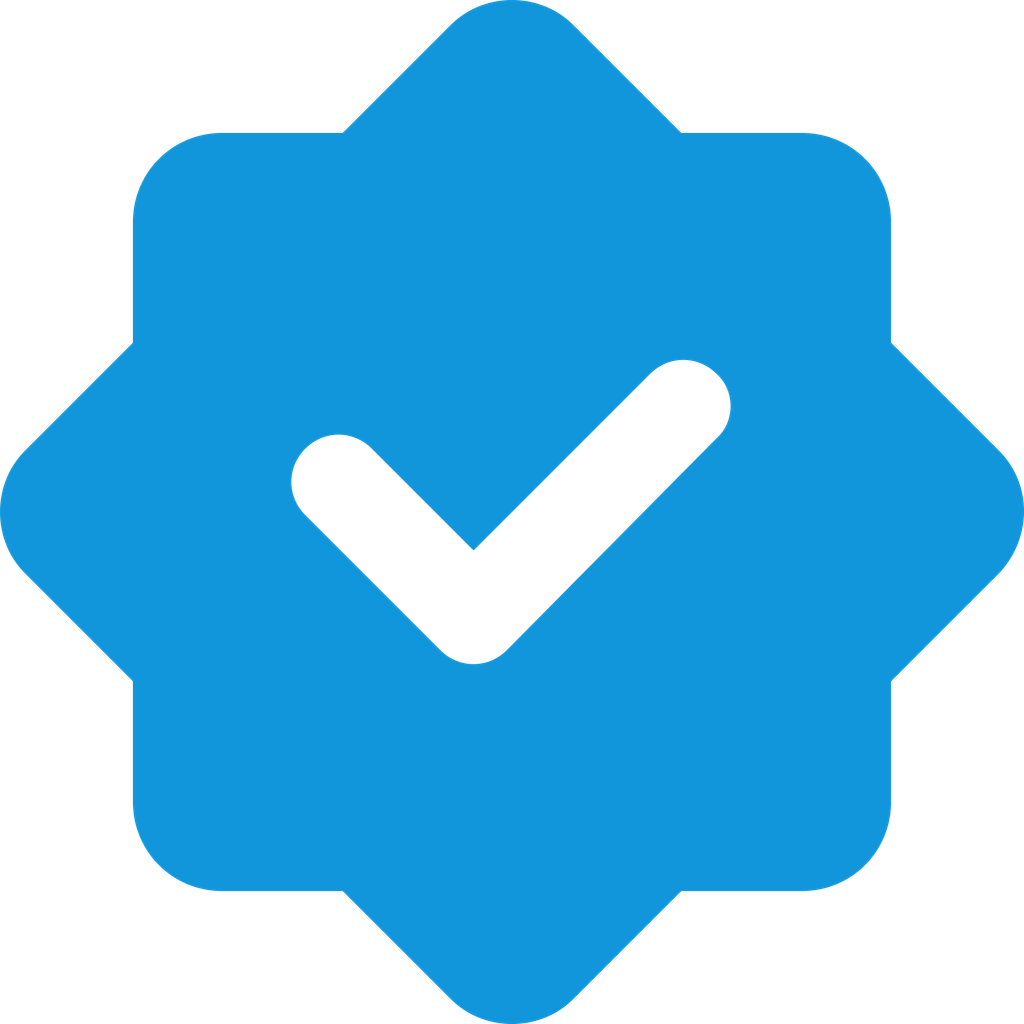}}%
  \endgroup
  \hspace{0.2em}}
\newlength{\captionmethodiconheight}
\DeclareRobustCommand{\captionmethodicon}[1]{%
  \begingroup
  \settoheight{\captionmethodiconheight}{X}%
  \raisebox{0pt}{\includegraphics[height=\captionmethodiconheight]{#1}}%
  \endgroup}
\DeclareRobustCommand{\captiontrainfreeicon}{\captionmethodicon{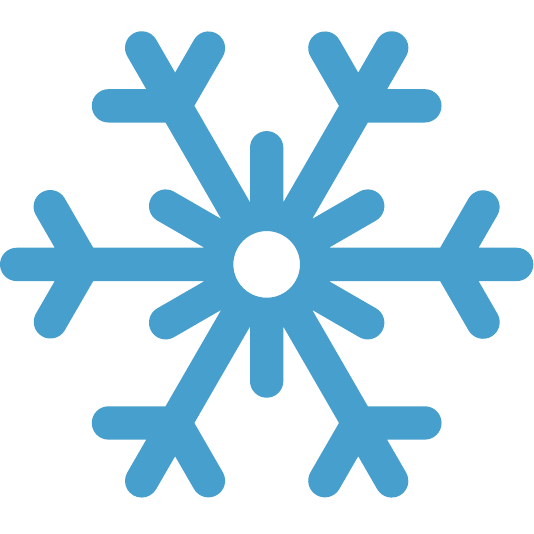}}
\DeclareRobustCommand{\captiontrainingicon}{\captionmethodicon{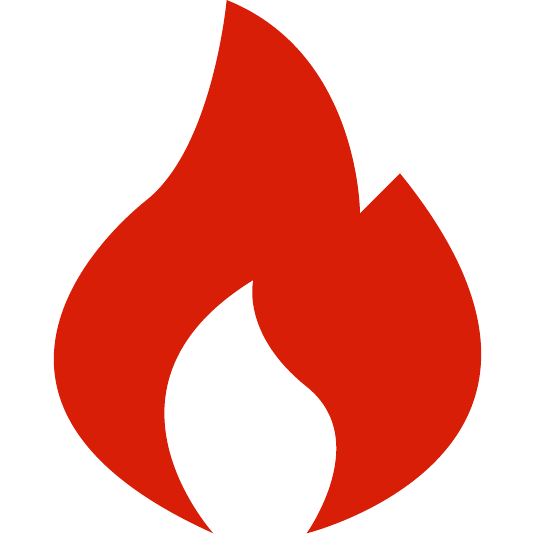}}
\newcommand{\tablemodelname}[1]{{\fontsize{6.5}{8}\selectfont\bfseries #1}}

\newcommand{\ind}{\mathbb{I}}

\begin{document}

\maketitle
\fancyhead{}
\renewcommand{\headrulewidth}{0pt}

\begin{abstract}

Large language models (LLMs) are increasingly deployed for natural-language logical reasoning, where the final answer is easy to check but the proof behind it is not.
In natural-language logical reasoning, intermediate conclusion should follow from its premises, and the resulting derivation should support the final answer.
Existing methods lack machine-checkable verification of intermediate conclusions and answer-supporting proof dependencies, so they may assign credit to invalid or answer-irrelevant steps.
We propose \textbf{\method{}}, a RL framework from formal verification that trains LLMs to construct verifiable proofs for natural-language logical reasoning.
\method{} admits a generated conclusion into the verified proof state only when the corresponding reasoning action satisfie proof obligations through UNSAT-based machine-checkable formal verification. 
\method{} also recovers the answer-supporting dependency closure to trace the proof structure of the final answer and align outcome credit with the proof dependencies.
Experiments demonstrate that \method{} improves answer accuracy across three logical reasoning benchmarks and four backbone models, and outperforms training-free agents and training-based methods in terms of reasoning-process verifiability.

\end{abstract}

\begin{figure}[H]
    \centering
    \includegraphics[height=0.255\linewidth]{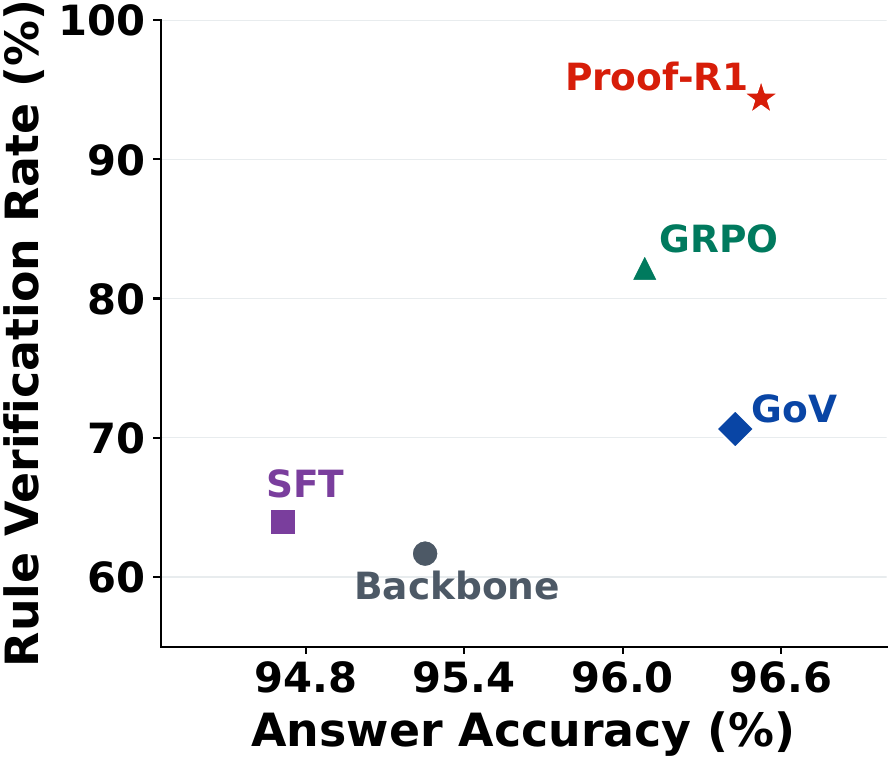}\hfill
    \includegraphics[height=0.255\linewidth]{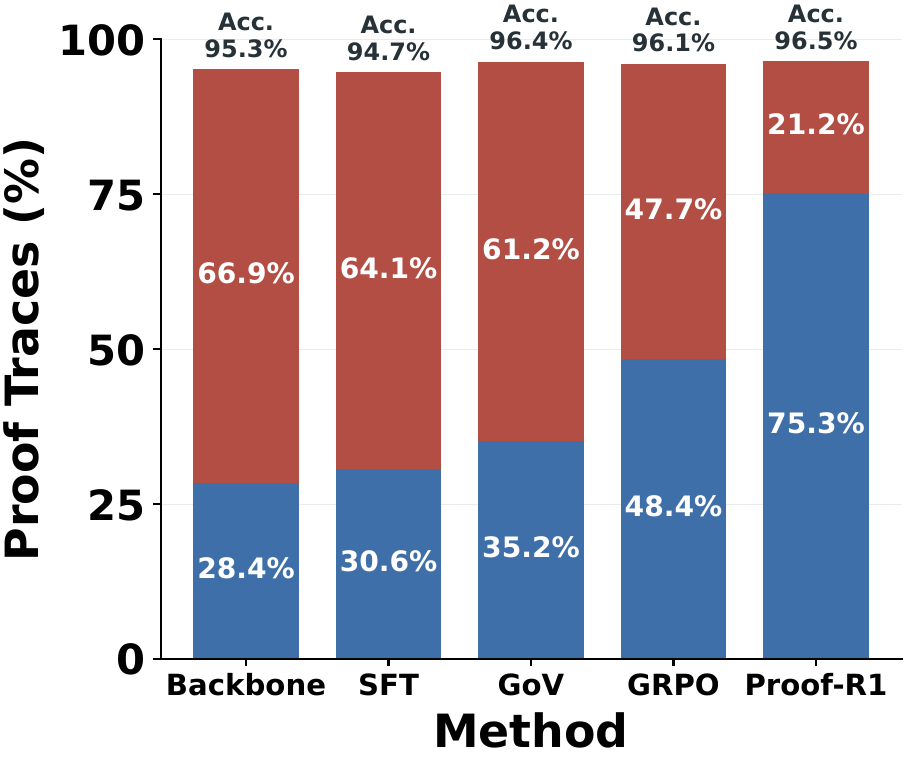}\hfill
    \includegraphics[height=0.255\linewidth]{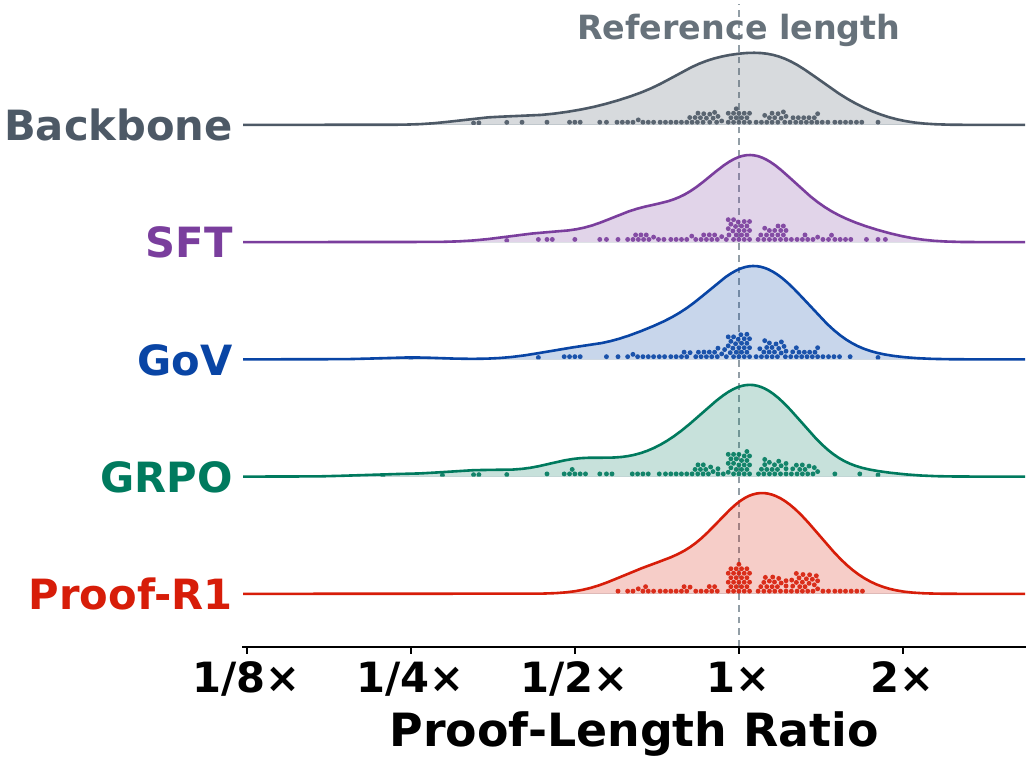}
    \vspace{-0.5em}
    \caption{
    Performance of \method{} on ProverQA.
    \textbf{Left}: Accuracy and step verification.
    The x-axis shows answer accuracy, while the y-axis shows RVR reflecting
    the logical correctness of generated reasoning under schema, semantic, and rule checks.
    \textbf{Middle}: Performance of different methods. The total height of each bar represents the proportion of correct answers, while the blue segment
    indicates the portion whose proof traces pass formal verification.
    \textbf{Right}: Generated-to-reference proof-length ratios across methods on hard subset of ProverQA.
    }
    \label{fig:insight}
\end{figure}

\section{Introduction}
\label{sec:introduction}

\begin{figure}[t]
    \centering
    \includegraphics[width=\linewidth]{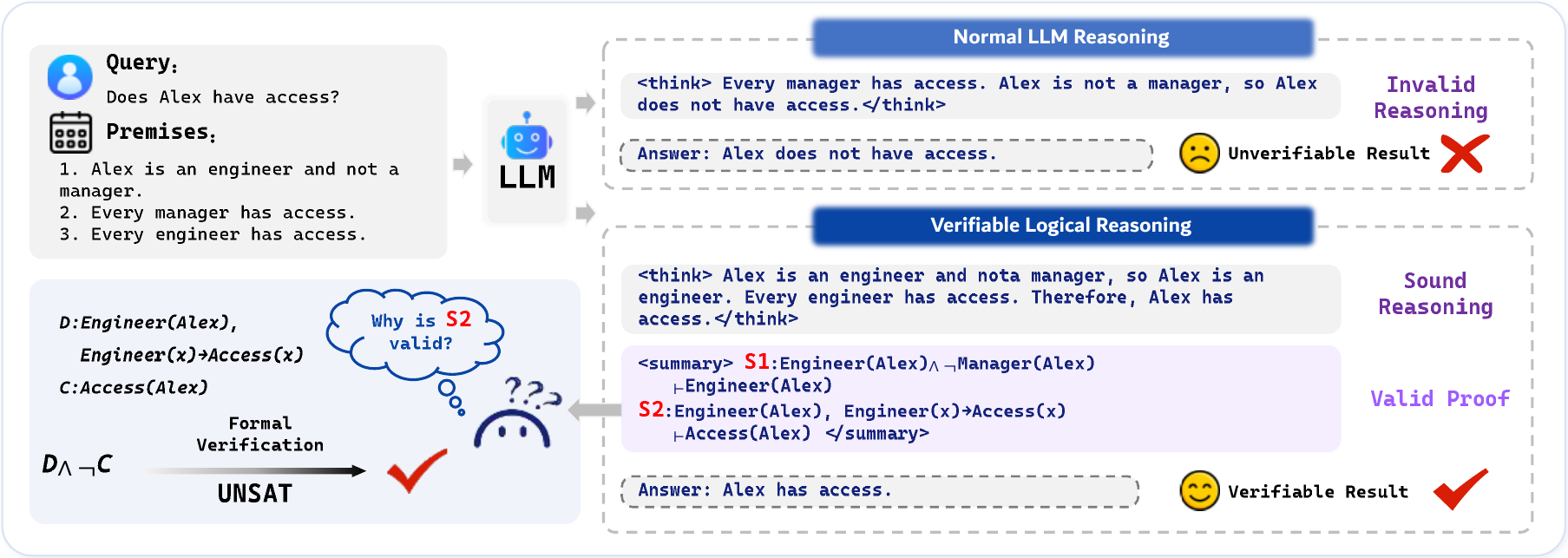}
    \setlength{\abovecaptionskip}{-6pt}
    \caption{Illustrative contrast between normal LLM reasoning and verifiable logical reasoning.
    The normal LLM reasoning incorrectly infers that ``Alex lacks access from not being
    a manager,'' committing the fallacy of denying the antecedent.
    The verifiable logical reasoning derives access from ``Alex's engineer status'' and exposes semantic proof obligation checked through formal verification.}

    \label{fig:reasoning_example}
\end{figure}

Recent advances in large language models (LLMs) have demonstrated strong
capabilities in multi-step reasoning and complex instruction following
\citep{wei2022chain,saparov2023ood,dubey2024llama3,he2024complex,
deepseekai2025r1,yang2025qwen3}.
As shown in Figure~\ref{fig:reasoning_example}, natural-language logical
reasoning asks whether a target conclusion follows from premises expressed
in both natural language and formal logic
\citep{clark2020ruletaker,tafjord2021proofwriter,han2024folio}.
A verifiable solution is a finite derivation in which every intermediate
conclusion follows from its declared premises alone under a valid rule, and the root conclusion establishes the target.

Post-training methods for natural-language logical reasoning commonly optimize answer-level objectives, with outcome-based reinforcement learning directly rewarding final-answer correctness \citep{shao2024deepseekmath,deepseekai2025r1,xie2025logicrl}.
In parallel, symbolic solvers and proof assistants have been used to inspect
model-generated reasoning, while formal verification feedback has been
incorporated into supervised learning, preference learning, and reinforcement
learning \citep{cao2025informal,liu2025safe,hubert2026alphaproof,fang2026gov}.

Despite their successes, each approach has inherent limitations.
Outcome rewards indicate whether a model reaches the correct answer but do
not identify whether its intermediate deductions are valid
\citep{shao2024deepseekmath,deepseekai2025r1,xie2025logicrl}.
Existing methods lack machine-checkable verification of intermediate conclusions and answer-supporting proof dependencies, so they may assign credit to invalid or answer-irrelevant steps
\citep{lightman2023verify,wang2024mathshepherd,xu2026logicreward,liu2025safe}.
Some methods introduce formal verification to inspect generated steps or construct rewards
\citep{liu2025safe,feng2025vericot,chen2026prosfi,fang2026gov}. However, a verification verdict alone determines neither which generated conclusions serve as trusted facts for subsequent inferences nor which verified inferences actually support the final answer.

This raises our central research question: \textit{\textbf{in natural-language logical reasoning, can formal verification do more
than provide a reward for a proof trace and instead define
the verifiable proof-construction process optimized by reinforcement learning?}}
To address this question, we propose \method{}, a RL
framework from formal verification that trains LLMs to construct verifiable proofs for
natural-language logical reasoning.
The policy represents a proof trace as a sequence of structured reasoning
actions, each specifying its dependencies, conclusion, and predicate-logic
rule.
For each candidate action, UNSAT-based \textit{machine-checkable formal verification} (\textbf{MCFV}) determines whether its conclusion follows from the declared dependencies and can therefore serve as a trusted dependency for subsequent actions.
Over the proof trace, \method{} constructs a candidate proof-certificate graph and recover the \textit{answer-supporting dependency closure} (\textbf{\ASDC}), thereby distinguishing answer-supporting actions from unused reasoning branches and making the proof basis of the final answer explicitly traceable.
\MCFV{} and \ASDC{} yield \textit{verification-aligned optimization}: formal verification no longer rewards, but determines which conclusions become trusted proof facts and which actions receive outcome credit during policy optimization.
Our contributions are as follows:
\begin{itemize}[leftmargin=1.5em,topsep=-1pt,partopsep=0pt]
    \item \textbf{Contribution 1} \textit{(Novelty of \method{})}: To address missing machine-checkable verification for intermediate conclusions and proof dependencies, we propose \method{}, a RL framework from formal verification that trains LLMs %
    for natural-language logical reasoning.
    \item \textbf{Contribution 2} \textit{(Performance of \method{})}: Across three logical reasoning benchmarks and four backbone models, \method{} improves answer accuracy and outperforms baselines in terms of reasoning-process verifiability.
    \item \textbf{Contribution 3} \textit{(Mechanistic Insights into \method{})}:
    We analyze \method{} at both the behavioral and optimization levels:
    rollout statistics provide empirical support for the design of \MCFV{},
    training dynamics reveal how the policy progressively learns to suppress
    invalid inferences through \ASDC{}, and parameter-gradient analysis explains how \method{} aligns optimization signals with verification evidence and proof dependencies.
\end{itemize}

\section{Related Work}
\label{sec:related_work}

\textbf{Natural-language logical reasoning task} is commonly formulated as determining
the entailment relation between a set of textual premises and a query.
In a verifiable solution, intermediate conclusions
should follow from their premises and the resulting derivation should support
the final answer
\citep{schwichtenberg2012proofs,tafjord2021proofwriter}.
Benchmarks such as RuleTaker and ProofWriter evaluate multi-step
deduction over textual facts and rules, with ProofWriter additionally
providing explicit proof targets
\citep{clark2020ruletaker,tafjord2021proofwriter}.
PrOntoQA uses synthetic problems with recoverable proof structures to evaluate
CoT reasoning and proof-planning abilities \citep{saparov2023greedy}.
FOLIO and ProverQA broaden benchmark coverage to more complex and diverse
first-order-logic problems through human annotation and prover-guided scalable
generation, respectively \citep{han2024folio,qi2025proverqa}.
Conventional approaches to these tasks include direct prediction\citep{clark2020ruletaker} and
supervised proof generation\citep{tafjord2021proofwriter}.

\textbf{Formal verification-augmented LLM reasoning} has emerged as a promising direction for
improving both performance and reliability, as LLM capabilities advance and their deployment expands.
Existing approaches differ in where formal verification enters the
reasoning, verification, and learning process.
At reasoning time, Logic-LM, LINC and LogicAgent use LLMs to formalize natural-language
problems and delegate deduction to symbolic solvers
\citep{pan2023logiclm,olausson2023linc,zhang2025logicagent}.
These reasoning-time pipelines improve problem solving, but do not provide a
policy-learning objective for constructing the verified proof itself.
A complementary line of work performs verification of generated
reasoning. Graph of Verification (GoV) organizes LLM-based verification in a
topologically ordered dependency graph with configurable granularity
\citep{fang2026gov}, while Safe and VeriCoT translate generated steps into
Lean statements or first-order arguments for tool-based checks
\citep{liu2025safe,feng2025vericot}.
For these methods, generation quality remains constrained by the backbone, and repeated verification calls can incur substantial token overhead.
Formal feedback has entered learning through several distinct mechanisms.
For natural-language reasoning, VeriCoT distills solver-validated traces into
supervised and preference-training data.\citep{feng2025vericot} LogicReward uses offline preference pairs generated by the policy and GPT-4o prior for supervised learning, while common theorem-proving reasoning datasets generally lack native pairwise preference annotations.
\citep{xu2026logicreward}.
PRoSFI generates structured intermediates and aggregates their formal checks
into a trajectory-level RL reward \citep{chen2026prosfi}, whereas Logic-RL
optimizes answer and format rewards without verifying intermediate deductions
\citep{xie2025logicrl}.
These methods primarily use formal verification to construct training signals, rather than allowing it to determine during policy optimization which generated conclusions become trusted proof facts and which actions receive outcome credit.

\section{Methodology}
\label{sec:method}

\begin{figure*}[t]
    \centering
    \includegraphics[width=\textwidth]{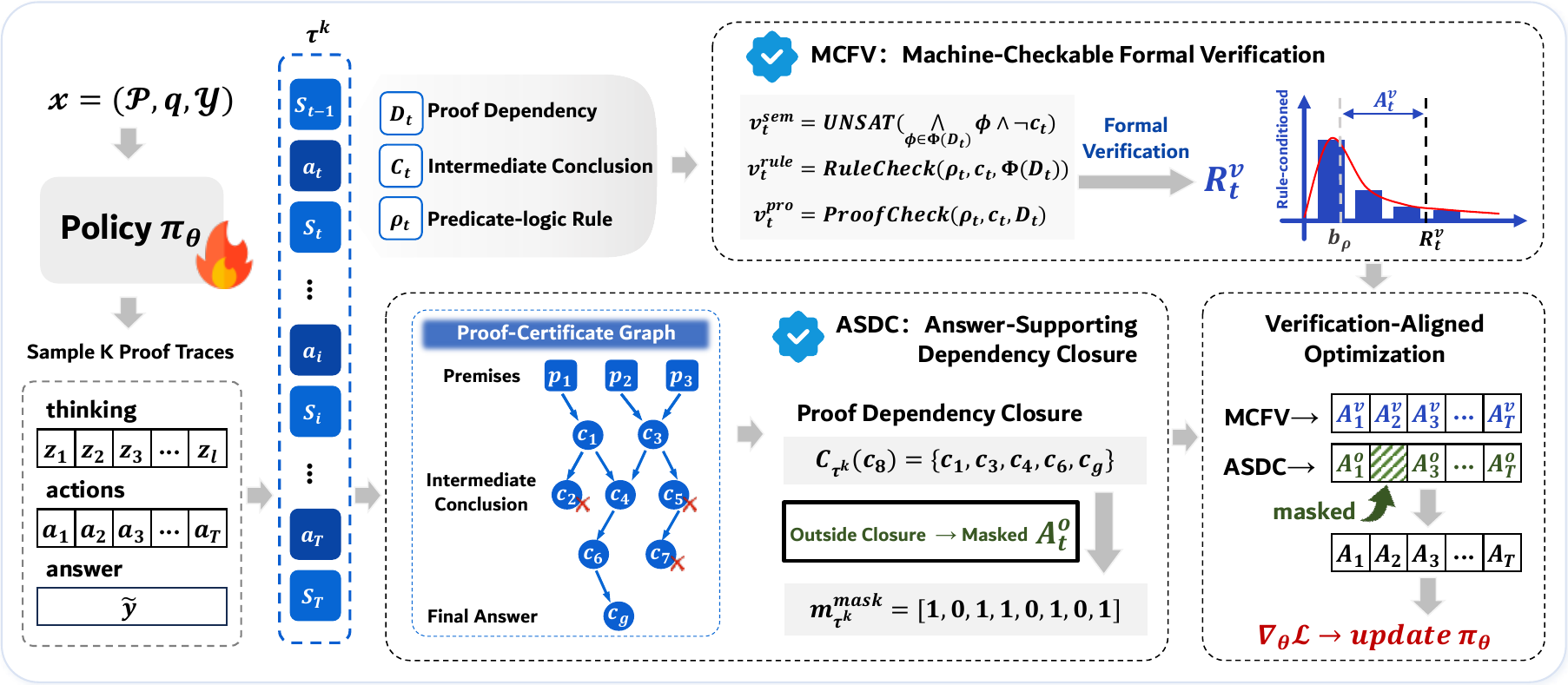}
    \caption{Framework of \method{}.
    \captionicon{}\MCFV{} formally verifies structured reasoning actions and maintains
    verified proof states.
    \captionicon{}\ASDC{} maintains the candidate proof-certificate graph and constructs the
    answer-supporting dependency closure.
    Verification-aligned optimization integrates the supervision signals from \MCFV{} and \ASDC{} into the policy update.}
    \label{fig:training}
\end{figure*}

\subsection{Problem formulation}
\label{sec:problem}

A natural-language logical reasoning instance is $x=(\mathcal{P},q,\mathcal{Y})$,
where $\mathcal{P}$ contains premises, $q$ is a query, and
$\mathcal{Y}=\{y_1,\ldots,y_N\}$ is a finite set of candidate answer propositions.
The premises in $\mathcal{P}$ are given both in natural language and as
formal logical expressions.
The task is to select a candidate entailed by $\mathcal{P}$ and construct a
proof of that candidate.
The reference answer $y^*\in\mathcal{Y}$ satisfies
\begin{equation}
\mathcal{P}\models y^*.
\label{eq:task_answer}
\end{equation}

Let $\mathcal{R}$ denote the trusted predicate-logic inference system. 
$\Pi$ is a finite dependency-structured derivation of $\tilde y$
from $\mathcal{P}$ under $\mathcal{R}$. A successful proof trace therefore satisfies:
\begin{equation}
    \tilde y=y^*,
    \qquad
    \mathcal{P}\vdash_{\mathcal{R}}^{\Pi}\tilde y
    \label{eq:certificate_judgment}
\end{equation}
Equation~\ref{eq:certificate_judgment} defines the proof-certificate
derivability judgment used in our framework.

\subsection{Structured proof generation}
\label{sec:overview}

As shown in Figure~\ref{fig:training}, \method{} links formal verification with proof-state transitions ($S_{t-1}\!\to S_t$) and proof dependencies ($D_t$) to train the policy to construct answer-supporting proofs.
For each problem $x$, the policy $\pi_\theta$ samples $K$ rollouts during
training.
The $k$-th rollout is
\begin{equation}
    \tau^k=(z^k,a_1^k,\ldots,a_{T_k}^k,\tilde y^k),
    \label{eq:rollout}
\end{equation}
where $z^k$ is free-form analysis, $T_k$ is the number of structured reasoning actions,
and $\tilde y^k$ is the selected candidate.

A checkable proof step should specify both its conclusion and the inference
claimed to justify it.
We represent an ordinary step at the granularity of a single predicate-logic
rule application:
\begin{equation}
    a_t=(D_t,c_t,\rho_t),
    \label{eq:action}
\end{equation}
where $D_t$ contains dependency identifiers, $c_t$ is the intermediate conclusion,
and $\rho_t$ is the declared predicate-logic rule.
Each action $a_t$ has a unique identifier $i_t$.
The terminal action $a_g$  matches the selected candidate, with conclusion $c_g=\tilde y$.
In subsequent descriptions of individual traces, we omit the rollout index $k$
to simplify the notation and make the equations easier to follow.

\subsection{Machine-checkable formal verification (\MCFV)}
\label{sec:action_verification}

A generated conclusion should pass verification before it can support a
subsequent verified inference.
As the verifier checks actions in generation order, it maintains two maps
from action identifiers to accepted conclusions: a semantically verified state
$\mathcal{S}^{\mathrm{sem}}_t$ and a stricter rule-verified state
$\mathcal{S}^{\mathrm{rule}}_t$.
Both maps are initially empty, while the original premises $\mathcal{P}$
remain available throughout verification.
Their pair $\mathcal{S}_t=(\mathcal{S}^{\mathrm{sem}}_t,
\mathcal{S}^{\mathrm{rule}}_t)$ is the aggregate proof state shown in
Figure~\ref{fig:training}.

Semantic verification checks whether the declared dependencies entail the proposed conclusion.
Before checking $a_t$, the verifier uses $\Phi_t^{\mathrm{sem}}(\cdot)$ to
resolve dependency identifiers into formulas against the original premises and
the semantic state constructed from preceding actions:
\begin{equation}
    \Phi_t^{\mathrm{sem}}(D_t)
    =\left\{
      \left(\mathcal{P}\cup\mathcal{S}^{\mathrm{sem}}_{t-1}\right)(d)
      \;\middle|\; d\in D_t
     \right\}.
    \label{eq:semantic_dependencies}
\end{equation}
Here, premises and accepted conclusions are indexed by their identifiers, so
$(\mathcal{P}\cup\mathcal{S})(d)$ denotes the formula named by dependency
$d$; an identifier absent from this union is rejected as an untrusted
dependency.
The semantic proof obligation is discharged when a satisfiability modulo
theories (SMT) solver returns UNSAT.
\begin{equation}
    \operatorname{VC}^{\mathrm{sem}}_t
    =\bigwedge_{\phi\in\Phi_t^{\mathrm{sem}}(D_t)}\phi\wedge\neg c_t,
    \qquad
    v_t^{\mathrm{sem}}
    =\operatorname{UNSAT}
      \left(\operatorname{VC}^{\mathrm{sem}}_t\right).
    \label{eq:semantic_verdict}
\end{equation}
A positive verdict establishes $\Phi_t^{\mathrm{sem}}(D_t)\models c_t$ and admits
the conclusion to the semantic state:
\begin{equation}
\mathcal{S}^{\mathrm{sem}}_t=
\begin{cases}
\mathcal{S}^{\mathrm{sem}}_{t-1}\cup\{i_t\mapsto c_t\},
&v_t^{\mathrm{schema}}\land v_t^{\mathrm{sem}},\\
\mathcal{S}^{\mathrm{sem}}_{t-1},&\text{otherwise}.
\end{cases}
\label{eq:semantic_state}
\end{equation}

Rule verification uses $\Phi_t^{\mathrm{rule}}(\cdot)$ to resolve dependency from the stricter rule-verified state:
\begin{equation}
    \Phi_t^{\mathrm{rule}}(D_t)
    =\left\{
      \left(\mathcal{P}\cup\mathcal{S}^{\mathrm{rule}}_{t-1}\right)(d)
      \;\middle|\; d\in D_t
    \right\},
    \qquad
    v_t^{\mathrm{rule}}
    =\operatorname{RuleCheck}
      \left(\rho_t,\Phi_t^{\mathrm{rule}}(D_t),c_t\right).
    \label{eq:rule_verdict}
\end{equation}

\textsc{RuleCheck} validates the rule's principal premises,
substitutions, and conclusion (Appendix~\ref{app:ontology}). A conclusion enters the rule-verified state only when its semantic and rule obligations are satisfied:
\begin{equation}
\mathcal{S}^{\mathrm{rule}}_t=
\begin{cases}
\mathcal{S}^{\mathrm{rule}}_{t-1}\cup\{i_t\mapsto c_t\},
&v_t^{\mathrm{schema}}\land v_t^{\mathrm{sem}}\land v_t^{\mathrm{rule}},\\
\mathcal{S}^{\mathrm{rule}}_{t-1},&\text{otherwise}.
\end{cases}
\label{eq:rule_state}
\end{equation}

\textsc{ProofCheck} checks whether an ordinary action makes nontrivial proof progress (Appendix~\ref{app:specification}):
\begin{equation}
    v_t^{\mathrm{pro}}=\operatorname{ProofCheck}(\rho_t,c_t,D_t).
    \label{eq:proofcheck_verdict}
\end{equation}

The verifier retains the four Boolean verdicts:
\begin{equation}
    v_t=(v_t^{\mathrm{schema}},v_t^{\mathrm{sem}},
         v_t^{\mathrm{rule}},v_t^{\mathrm{pro}})
         \in\{\mathrm{true},\mathrm{false}\}^{4}.
    \label{eq:structured_verdict}
\end{equation}

During trace verification, each state is therefore both the output at step
$t$ and the verifier memory used to resolve dependencies at step $t+1$;
downstream optimization consumes the resulting verdicts.
The structured verdict $v_t^k$ of $a_t^k$ is mapped to a numerical verification signal:
\begin{equation}
R_{k,t}^{v}=
\begin{cases}
0.0, & \neg v_t^{\mathrm{schema},k},\\
0.1, & v_t^{\mathrm{schema},k}\land\neg v_t^{\mathrm{sem},k},\\
0.3, & v_t^{\mathrm{schema},k}\land v_t^{\mathrm{sem},k}
       \land\bigl(\neg v_t^{\mathrm{rule},k}\lor\neg v_t^{\mathrm{pro},k}\bigr),\\
1.0, & v_t^{\mathrm{schema},k}\land v_t^{\mathrm{sem},k}
       \land v_t^{\mathrm{rule},k}\land v_t^{\mathrm{pro},k}.
\end{cases}
\label{eq:verification_signal}
\end{equation}

We center each verification signal using a rule-conditioned centering value
estimated from the recent verification outcomes of actions that invoke the
same predicate-logic rule.
Let $\operatorname{EMA}_\rho^{(u-1)}$ denote the exponential moving average
for rule $\rho$ before update $u$.
The clipped rule-conditioned centering value and action-verification advantage are:
\begin{align}
b_\rho^{(u-1)}
=\operatorname{clip}(\operatorname{EMA}_\rho^{(u-1)},b_{\min},b_{\max}),
\qquad
A_{k,t}^{v}=R_{k,t}^{v}-b_{\rho_t^k}^{(u-1)}.
\label{eq:verification_advantage}
\end{align}
For each rule observed in the current batch, let $\mathcal{B}_\rho^{(u)}$
contain the actions assigned to that rule.
After computing the current advantages, we update:
\begin{align}
\overline{R}_\rho^{v,(u)}=\frac{1}{|\mathcal{B}_\rho^{(u)}|}
  \sum_{a_t^k\in\mathcal{B}_\rho^{(u)}}R_{k,t}^{v},
\qquad
\operatorname{EMA}_\rho^{(u)}=\beta\operatorname{EMA}_\rho^{(u-1)}
 +(1-\beta)\overline{R}_\rho^{v,(u)}.
\label{eq:ema_update}
\end{align}

\subsection{Answer-supporting dependency closure (\ASDC)}
\label{sec:proof_construction}

A verified conclusion need not participate in the proof of the selected answer.
To identify the declared answer derivation, we construct a verifier-annotated
candidate proof-certificate graph $G_\tau=(V_\tau,E_\tau)$. Writing the
original premises as $\mathcal{P}=\{p_j\}_{j=1}^{M}$, its nodes and edges are
\begin{equation}
\begin{aligned}
V_\tau
=\mathcal{P}\cup\{c_t\}_{t=1}^{T},
\qquad 
E_\tau=\left\{(u,c_t)\;\middle|\;
u\in\mathcal{P}\cup\{c_j\}_{j<t},\ 
\operatorname{id}(u)\in D_t\right\}.
\end{aligned}
\label{eq:certificate_graph}
\end{equation}
Here $\operatorname{id}(p_j)$ is the identifier of an original premise and
$\operatorname{id}(c_j)=i_j$ is the identifier of the action that generated
$c_j$.
The answer-supporting dependency closure rooted at the final conclusion $c_g$
is the closed ancestor set:
\begin{equation}
\begin{aligned}
    \mathcal{C}_\tau(c_g)
    &:= \operatorname{Anc}_{G_\tau}(c_g)\cup\{c_g\}=\{u\in V_\tau:u\leadsto_{G_\tau}c_g\}\cup\{c_g\},
\end{aligned}
    \label{eq:dependency_closure}
\end{equation}
where $\operatorname{Anc}_{G_\tau}(c_g)$ denotes the strict ancestors of
$c_g$, and $\leadsto_{G_\tau}$ denotes directed reachability in $G_\tau$.
We compute the closure by traversing dependencies backward from $c_g$ until
all ancestor conclusions and premises have been included.
The proof-certificate graph for the selected answer is the subgraph induced by this
closure:
\begin{equation}
    \Pi_\tau
    =G_\tau\!\left[\mathcal{C}_\tau(c_g)\right].
    \label{eq:proof_certificate}
\end{equation}
Thus, $\Pi_\tau$ contains exactly the premise and generated-conclusion
nodes declared to support the final answer, together with their dependency
relations.

A correct answer does not make every action in its response part of the
answer's proof.
The closure mask maps conclusion membership back to the action that generated
each conclusion:
\begin{equation}
m_{k,t}=\ind[c_t^k\in\mathcal{C}_{\tau^k}(c_g^k)],
\qquad 
\mathbf{m}^k=(m_{k,1},\ldots,m_{k,T_k}).
\end{equation}

For rollout $k$, the outcome signal is
$R_k^{o}=\ind[\tilde y^k=y^*]$.
We normalize this signal over eligible rollouts for the same problem:
\begin{equation}
\widehat A_k^{o}
=\frac{R_k^{o}-\mu_x}{\sigma_x+\epsilon}.
\label{eq:outcome_advantage}
\end{equation}
Here $\mu_x$ and $\sigma_x$ are the group mean and standard deviation.
The mask restricts outcome credit to actions that support the answer:
\begin{equation}
A_{k,t}^{o}=m_{k,t}\widehat A_k^{o}.
\label{eq:proof_advantage}
\end{equation}
Consequently, an action receives no outcome advantage when its generated
conclusion lies outside the closure, even when the final answer is correct.

\subsection{Verification-aligned optimization}
\label{sec:aligned_rl}

Verification-aligned optimization combines the action-verification advantage
from \MCFV{} with the closure-masked outcome advantage from \ASDC{}, and
aligns the resulting action-level advantages with the exact token spans of
structured actions:
\begin{equation}
A_{k,t}=\lambda_v A_{k,t}^{v}+\lambda_o A_{k,t}^{o},
\qquad
\mathbf{A}^{\mathrm{tok},k}=M^k\mathbf{A}^{k}.
\label{eq:advantage_projection}
\end{equation}
Here $\mathbf{A}^k$ collects the action-level advantages, and $M^k$ is the
normalized action-to-token alignment matrix that maps each action advantage
to its exact token span.
We then optimize the policy with a KL-regularized PPO-style clipped objective
\citep{schulman2017ppo,shao2024deepseekmath}.
And we provide the construction of $M^k$, full objective, and sampled KL estimator (Appendix~\ref{app:policy_implementation}).

\section{Experiments}
\label{sec:experiments}

\subsection{Experimental Setup}

\paragraph{Datasets and models.}
We train and evaluate \method{} on ProverQA \citep{qi2025proverqa}.
We evaluate two backbones without a native thinking mode, Qwen2.5-7B-Instruct
\citep{yang2024qwen25} and Llama-3.1-8B \citep{dubey2024llama3}, and two
backbones with native thinking support, Qwen3-8B \citep{yang2025qwen3} and
GLM-Z1-9B-0414 \citep{thudm2025glmz1}.
We further assess cross-benchmark generalization on FOLIO
\citep{han2024folio} and ProofWriter \citep{tafjord2021proofwriter}.
Notably, the evaluated FOLIO subset contains valid derivations that require a
strictly broader inference system,
$\mathcal{R}_{\mathrm{FOLIO}}\supset\mathcal{R}_{\mathrm{train}}$, with universal
and existential quantifier rule families (Appendix~\ref{app:quantifier_coverage}).

\paragraph{Baselines.}
Comparisons of training-free agents include LogicAgent \citep{zhang2025logicagent} and GoV \citep{fang2026gov}.
LogicAgent uses an LLM to reason over natural-language problems and delegates
deduction to symbolic solvers.
GoV organizes LLM-based verification in a topologically ordered dependency
graph with configurable granularity.
The training-based methods comprise SFT, GRPO \citep{shao2024deepseekmath}, and PRoSFI \citep{chen2026prosfi}.
PRoSFI is an RL method for natural-language logical reasoning.

\paragraph{Evaluation metrics.}
We report Avg@3 and Pass@3 for answer correctness.
We assess proof quality using \textit{Formal Verification Rate} (FVR)
\citep{xu2026logicreward}, \textit{Rule Verification Rate} (RVR)
\citep{zhou2025rulearena}, and \textit{Reasoning Granularity Deviation} (RGD)
\citep{gu2026proofoptimizer}(Appendix~\ref{app:evaluation_metrics}).\footnote{Because FOLIO provides no reference proofs, whereas ProofWriter provides reference traces generated under a Horn inference system distinct from $\mathcal{R}_{\mathrm{train}}$, RGD is not used as a common metric in Tables~\ref{tab:cross_benchmark_results} and~\ref{tab:qwen25_cross_dataset}.}

\newcommand{\meanstdcell}[2]{\mbox{\makebox[2.5em][r]{#1}\hspace{0.08em}\scalebox{0.60}{\textcolor{gray}{#2}}}}
\newcommand{\trainfreeicon}{\raisebox{-0.15ex}{\includegraphics[height=0.9em]{figures/icons/train_free_snowflake.pdf}}\hspace{0.25em}}
\newcommand{\trainingicon}{\raisebox{-0.15ex}{\includegraphics[height=0.9em]{figures/icons/training_based_flame.pdf}}\hspace{0.25em}}
\begin{table}[!t]
    \setlength{\belowcaptionskip}{2.5pt}
    \caption{Results on ProverQA, reported as means followed by standard deviations in gray. \captiontrainfreeicon{} and \captiontrainingicon{} denote training-free and training-based methods, respectively. \textbf{Bold} and \underline{underline} denote the best and second-best means within each backbone model. ProverQA-Hard denotes the hard split of ProverQA, whereas ProverQA-Overall denotes the evaluation set across all difficulty levels.}
    \label{tab:main_results}
    \centering
    \fontsize{8}{8.5}\selectfont
    \setlength{\tabcolsep}{1pt}
    \renewcommand{\arraystretch}{1.2}
    \resizebox{\linewidth}{!}{%
    \begin{tabular}{@{}cl*{5}{c}@{\hspace{5pt}}*{5}{c}@{}}
        \toprule
        \multicolumn{1}{c}{\multirow{2}{*}{\textbf{Model}}} & \multicolumn{1}{c}{\multirow{2}{*}{\textbf{Method}}} & \multicolumn{5}{c}{\textbf{ProverQA-Overall}}
        & \multicolumn{5}{c}{\textbf{ProverQA-Hard}} \\
        \cmidrule(lr){3-7}\cmidrule(lr){8-12}
        & & Avg@3 $\uparrow$ & Pass@3 $\uparrow$ & FVR $\uparrow$ & RVR $\uparrow$ & RGD $\downarrow$
        & Avg@3 $\uparrow$ & Pass@3 $\uparrow$ & FVR $\uparrow$ & RVR $\uparrow$ & RGD $\downarrow$ \\
        \midrule
        \multirow{7}{*}[-2pt]{\rotatebox[origin=c]{90}{\tablemodelname{Qwen2.5-7B-Instruct}}} & \trainfreeicon Backbone & \meanstdcell{44.79}{0.25} & \meanstdcell{68.43}{1.78} & \meanstdcell{67.64}{0.68} & \meanstdcell{18.91}{1.09} & \meanstdcell{0.88}{0.017} & \meanstdcell{27.14}{0.58} & \meanstdcell{51.76}{3.54} & \meanstdcell{56.59}{1.08} & \meanstdcell{9.84}{0.43} & \meanstdcell{1.36}{0.022} \\
        & \trainfreeicon LogicAgent & \meanstdcell{59.28}{0.56} & \meanstdcell{81.20}{1.49} & \meanstdcell{54.39}{1.02} & \meanstdcell{\underline{30.44}}{0.55} & \meanstdcell{0.93}{0.009} & \meanstdcell{\underline{40.03}}{0.44} & \meanstdcell{\underline{65.83}}{3.37} & \meanstdcell{38.11}{1.42} & \meanstdcell{\underline{15.72}}{1.67} & \meanstdcell{1.51}{0.008} \\
        & \trainfreeicon GoV & \meanstdcell{47.23}{0.82} & \meanstdcell{70.63}{1.75} & \meanstdcell{65.28}{0.54} & \meanstdcell{23.38}{0.45} & \meanstdcell{0.87}{0.010} & \meanstdcell{26.80}{1.60} & \meanstdcell{51.76}{3.54} & \meanstdcell{54.50}{1.09} & \meanstdcell{10.28}{1.07} & \meanstdcell{1.34}{0.010} \\
        \addlinespace[2pt]
        & \trainingicon SFT & \meanstdcell{47.04}{0.73} & \meanstdcell{70.78}{1.75} & \meanstdcell{66.39}{0.55} & \meanstdcell{19.94}{0.86} & \meanstdcell{\underline{0.80}}{0.018} & \meanstdcell{30.82}{1.17} & \meanstdcell{53.77}{3.54} & \meanstdcell{\underline{57.59}}{0.88} & \meanstdcell{9.49}{0.83} & \meanstdcell{\underline{1.15}}{0.028} \\
        \addlinespace[2pt]
        & \trainingicon GRPO & \meanstdcell{47.28}{1.33} & \meanstdcell{76.36}{1.63} & \meanstdcell{\underline{68.09}}{0.84} & \meanstdcell{12.50}{0.66} & \meanstdcell{1.02}{0.013} & \meanstdcell{23.45}{1.17} & \meanstdcell{49.75}{3.56} & \meanstdcell{48.85}{1.72} & \meanstdcell{6.03}{0.51} & \meanstdcell{1.63}{0.027} \\
        & \trainingicon PRoSFI & \meanstdcell{\underline{59.57}}{0.68} & \meanstdcell{\underline{84.21}}{1.04} & \meanstdcell{62.59}{0.55} & \meanstdcell{15.22}{0.97} & \meanstdcell{0.81}{0.009} & \meanstdcell{38.61}{0.77} & \meanstdcell{65.58}{2.76} & \meanstdcell{47.84}{1.18} & \meanstdcell{8.34}{0.80} & \meanstdcell{1.18}{0.013} \\
        & \trainingicon\textbf{\method{}} & \meanstdcell{\textbf{62.80}}{0.56} & \meanstdcell{\textbf{87.08}}{1.29} & \meanstdcell{\textbf{70.17}}{0.28} & \meanstdcell{\textbf{41.19}}{0.44} & \meanstdcell{\textbf{0.70}}{0.018} & \meanstdcell{\textbf{44.89}}{0.60} & \meanstdcell{\textbf{76.88}}{2.99} & \meanstdcell{\textbf{58.74}}{0.26} & \meanstdcell{\textbf{19.38}}{0.75} & \meanstdcell{\textbf{1.05}}{0.033} \\
        \midrule
        \multirow{7}{*}[-2pt]{\rotatebox[origin=c]{90}{\tablemodelname{Qwen3-8B}}} & \trainfreeicon Backbone & \meanstdcell{95.25}{0.53} & \meanstdcell{98.83}{0.21} & \meanstdcell{81.82}{0.36} & \meanstdcell{61.68}{0.11} & \meanstdcell{0.33}{0.011} & \meanstdcell{91.12}{1.31} & \meanstdcell{96.50}{0.50} & \meanstdcell{81.12}{0.91} & \meanstdcell{56.40}{1.03} & \meanstdcell{0.40}{0.026} \\
        & \trainfreeicon LogicAgent & \meanstdcell{\textbf{96.87}}{0.42} & \meanstdcell{\underline{99.12}}{0.36} & \meanstdcell{93.88}{0.19} & \meanstdcell{79.70}{0.55} & \meanstdcell{0.33}{0.009} & \meanstdcell{91.29}{1.57} & \meanstdcell{97.49}{1.11} & \meanstdcell{90.39}{0.52} & \meanstdcell{67.14}{1.71} & \meanstdcell{0.47}{0.025} \\
        & \trainfreeicon GoV & \meanstdcell{96.43}{0.05} & \meanstdcell{98.53}{0.46} & \meanstdcell{89.05}{0.61} & \meanstdcell{70.63}{0.63} & \meanstdcell{\underline{0.29}}{0.006} & \meanstdcell{\textbf{92.96}}{0.77} & \meanstdcell{97.49}{1.10} & \meanstdcell{86.04}{0.57} & \meanstdcell{61.69}{0.28} & \meanstdcell{0.38}{0.017} \\
        \addlinespace[2pt]
        & \trainingicon SFT & \meanstdcell{94.71}{0.37} & \meanstdcell{\underline{99.12}}{0.36} & \meanstdcell{83.00}{0.06} & \meanstdcell{63.94}{1.01} & \meanstdcell{0.33}{0.008} & \meanstdcell{89.11}{0.67} & \meanstdcell{\underline{97.99}}{0.99} & \meanstdcell{78.99}{1.44} & \meanstdcell{55.97}{1.93} & \meanstdcell{0.41}{0.019} \\
        \addlinespace[2pt]
        & \trainingicon GRPO & \meanstdcell{96.08}{0.44} & \meanstdcell{\textbf{99.27}}{0.33} & \meanstdcell{94.22}{0.32} & \meanstdcell{82.23}{0.55} & \meanstdcell{0.31}{0.007} & \meanstdcell{90.28}{1.47} & \meanstdcell{\underline{97.99}}{0.99} & \meanstdcell{91.06}{0.32} & \meanstdcell{73.26}{0.70} & \meanstdcell{0.39}{0.026} \\
        & \trainingicon PRoSFI & \meanstdcell{96.23}{0.13} & \meanstdcell{98.83}{0.33} & \meanstdcell{\underline{95.13}}{0.26} & \meanstdcell{\underline{89.28}}{0.38} & \meanstdcell{\underline{0.29}}{0.006} & \meanstdcell{92.13}{0.50} & \meanstdcell{\underline{97.99}}{0.99} & \meanstdcell{\underline{92.88}}{0.53} & \meanstdcell{\underline{84.92}}{0.69} & \meanstdcell{\underline{0.37}}{0.011} \\
        & \trainingicon\textbf{\method{}} & \meanstdcell{\underline{96.52}}{0.26} & \meanstdcell{\textbf{99.27}}{0.33} & \meanstdcell{\textbf{96.56}}{0.15} & \meanstdcell{\textbf{94.36}}{0.08} & \meanstdcell{\textbf{0.28}}{0.005} & \meanstdcell{\underline{92.63}}{1.10} & \meanstdcell{\textbf{99.00}}{0.71} & \meanstdcell{\textbf{93.55}}{0.32} & \meanstdcell{\textbf{90.22}}{0.26} & \meanstdcell{\textbf{0.36}}{0.002} \\
        \bottomrule
    \end{tabular}}
\end{table}

\subsection{Results}
\label{sec:results}

Table~\ref{tab:main_results} presents the main results of \method{} on ProverQA.
On Qwen2.5-7B-Instruct, \method{} improves Avg@3, Pass@3, FVR, and RVR over
the strongest Overall-split baseline for each metric by 3.23, 2.87, 2.08,
and 10.75 points, respectively, while reducing RGD from 0.80 to 0.70.
On the more challenging Hard split, it improves Avg@3, Pass@3, FVR, and RVR
by 4.86, 11.05, 1.15, and 3.66 points, respectively, while reducing RGD from
1.15 to 1.05.
On Qwen3-8B, answer accuracy is near saturation. The training-free agents
LogicAgent and GoV attain slightly higher Avg@3 than \method{} on the Overall
and Hard splits. Both methods obtain these marginal gains through
multiple model calls for multi-stage reasoning, verification, and answer
selection, which substantially increases inference-time token consumption
(Appendix~\ref{app:computational_cost}).
On the Overall-split, \method{} achieves an FVR of 96.56 and an RVR of 94.36,
improving over the strongest baselines by 1.43 and 5.08 points,
and obtains the lowest RGD of 0.28.
The results indicate that \emph{our method improves the formal-verification quality of proof steps while preserving answer accuracy}.

\subsection{Ablations}
\label{sec:verification_controls}

\begin{wraptable}{r}{0.51\linewidth}
\vspace{-2\baselineskip}

    \caption{Ablation results on ProverQA-Hard.}
    \label{tab:ablations}
    \centering
    \fontsize{8}{8.5}\selectfont
    \setlength{\tabcolsep}{0.6pt}
    \renewcommand{\arraystretch}{1.20}
    \begin{tabular*}{\linewidth}{@{\extracolsep{\fill}}clccccc@{}}
        \toprule
        \multicolumn{1}{c}{\textbf{Model}} & \multicolumn{1}{c}{\textbf{Variant}} & Avg@3 $\uparrow$ & Pass@3 $\uparrow$ & FVR $\uparrow$ & RVR $\uparrow$ & RGD $\downarrow$ \\
        \midrule
        \multirow{3}{*}{\rotatebox[origin=c]{90}{\tablemodelname{\shortstack{Qwen2.5\\7B-Instruct}}}}
        & w/o \MCFV{} & 37.86 & 68.34 & 40.16 & 11.65 & 1.08 \\
        & w/o \ASDC{} & 41.37 & 71.86 & 44.18 & 16.69 & 1.34 \\
        & \textbf{\method{}} & \textbf{44.89} & \textbf{76.88} & \textbf{58.74} & \textbf{19.38} & \textbf{1.05} \\
        \midrule
        \multirow{3}{*}{\rotatebox[origin=c]{90}{\tablemodelname{Qwen3-8B}}}
        & w/o \MCFV{} & 90.28 & 98.49 & 82.52 & 57.73 & 0.37 \\
        & w/o \ASDC{} & 89.61 & 97.99 & \textbf{93.55} & 85.71 & 0.40 \\
        & \textbf{\method{}} & \textbf{92.63} & \textbf{99.00} & \textbf{93.55} & \textbf{90.22} & \textbf{0.36} \\
        \bottomrule
    \end{tabular*}

\vspace{-0.5\baselineskip}
\end{wraptable}

Table~\ref{tab:ablations} examines the roles of \MCFV{} and \ASDC{} in
\method{}.
Across both backbone models on ProverQA-Hard, the full method achieves the best result on every metric, outperforming both ablated variants.
The w/o \MCFV{} variant obtains a lower RGD than w/o \ASDC{}
on both backbones (1.08 vs.\ 1.34 and 0.37 vs.\ 0.40), indicating that \ASDC{}
guides the policy toward proof traces whose granularity more closely matches
the reference proofs.
Conversely, the w/o \ASDC{} variant achieves higher FVR  and
RVR than w/o \MCFV{} on both backbones, showing that \MCFV{} directly improves the semantic validity of generated proof steps.

Particularly, the relative answer accuracy of the two ablations reverses across
backbones.
On Qwen2.5-7B-Instruct, w/o \ASDC{} outperforms w/o \MCFV{} in Avg@3 (41.37 vs.\ 37.86), whereas on Qwen3-8B, w/o \MCFV{} outperforms w/o \ASDC{} (90.28 vs.\ 89.61).
This pattern may suggest that \emph{weaker models benefit more from improving the validity of proof steps, whereas stronger models benefit increasingly from distinguishing answer-supporting actions with irrelevant branches.}

\begin{table}[!t]
    \begin{minipage}[t]{0.515\linewidth}
        \vspace{0pt}
        \caption{Performance of \method{} trained with backbones from additional model families on ProverQA-Overall.\strut}
\label{tab:backbone_results}
\centering
\fontsize{8}{8.5}\selectfont
\setlength{\tabcolsep}{0.6pt}
\renewcommand{\arraystretch}{1.20}
\begin{tabular*}{\linewidth}{@{\extracolsep{\fill}}c@{\hspace{3pt}}l*{5}{c}@{}}
    \toprule
    \multicolumn{1}{c}{\textbf{Model}} & \multicolumn{1}{c}{\textbf{Method}} & Avg@3 $\uparrow$ & Pass@3 $\uparrow$ & FVR $\uparrow$ & RVR $\uparrow$ & RGD $\downarrow$ \\
    \midrule
    \multirow{3}{*}[-1pt]{\rotatebox[origin=c]{90}{\tablemodelname{\shortstack{Llama-3.1\\8B}}}} & Backbone & 46.30 & 76.54 & 50.11 & 9.20 & 0.94 \\
    & SFT & 44.65 & 75.19 & 46.23 & 9.61 & 0.92 \\
    & \textbf{\method{}} & \textbf{53.95} & \textbf{82.84} & \textbf{54.72} & \textbf{23.02} & \textbf{0.91} \\
    \midrule
    \multirow{3}{*}[-1pt]{\rotatebox[origin=c]{90}{\tablemodelname{\shortstack{GLM-Z1\\9B-0414}}}} & Backbone & 71.76 & 91.92 & 82.64 & 52.47 & 0.70 \\
    & SFT & 74.40 & 96.48 & 83.03 & 57.23 & 0.65 \\
    & \textbf{\method{}} & \textbf{89.33} & \textbf{99.27} & \textbf{95.01} & \textbf{87.26} & \textbf{0.42} \\
    \bottomrule
\end{tabular*}

    \end{minipage}\hfill
    \begin{minipage}[t]{0.465\linewidth}
        \vspace{0pt}
        \caption{Cross-dataset generalization of \method{} trained on ProverQA, tested on FOLIO and ProofWriter.\strut}
\label{tab:cross_benchmark_results}
\centering
\fontsize{8}{8.5}\selectfont
\setlength{\tabcolsep}{0.6pt}
\renewcommand{\arraystretch}{1.20}
\begin{tabular*}{\linewidth}{@{\extracolsep{\fill}}c@{\hspace{3pt}}l*{4}{c}@{}}
    \toprule
    \multicolumn{1}{c}{\textbf{Dataset}} & \multicolumn{1}{c}{\textbf{Method}} & Avg@3 $\uparrow$ & Pass@3 $\uparrow$ & FVR $\uparrow$ & RVR $\uparrow$ \\
    \midrule
    \multirow{3}{*}[-1pt]{\rotatebox[origin=c]{90}{\tablemodelname{FOLIO}}} & Backbone & 88.89 & \textbf{98.04} & 64.46 & 38.87 \\
    & SFT & 85.62 & \textbf{98.04} & 63.22 & 41.03 \\
    & \textbf{\method{}} & \textbf{90.85} & \textbf{98.04} & \textbf{82.58} & \textbf{59.70} \\
    \midrule
    \multirow{3}{*}[-1pt]{\rotatebox[origin=c]{90}{\tablemodelname{\shortstack{Proof\\Writer}}}} & Backbone & 80.89 & 90.67 & 77.07 & 51.78 \\
    & SFT & 82.22 & 92.67 & 64.72 & 40.97 \\
    & \textbf{\method{}} & \textbf{94.89} & \textbf{98.00} & \textbf{97.57} & \textbf{85.46} \\
    \bottomrule
\end{tabular*}

    \end{minipage}
\end{table}

\begin{figure}[!t]
    \centering
    \begin{minipage}[t]{0.32\linewidth}
        \centering
        \includegraphics[width=\linewidth]{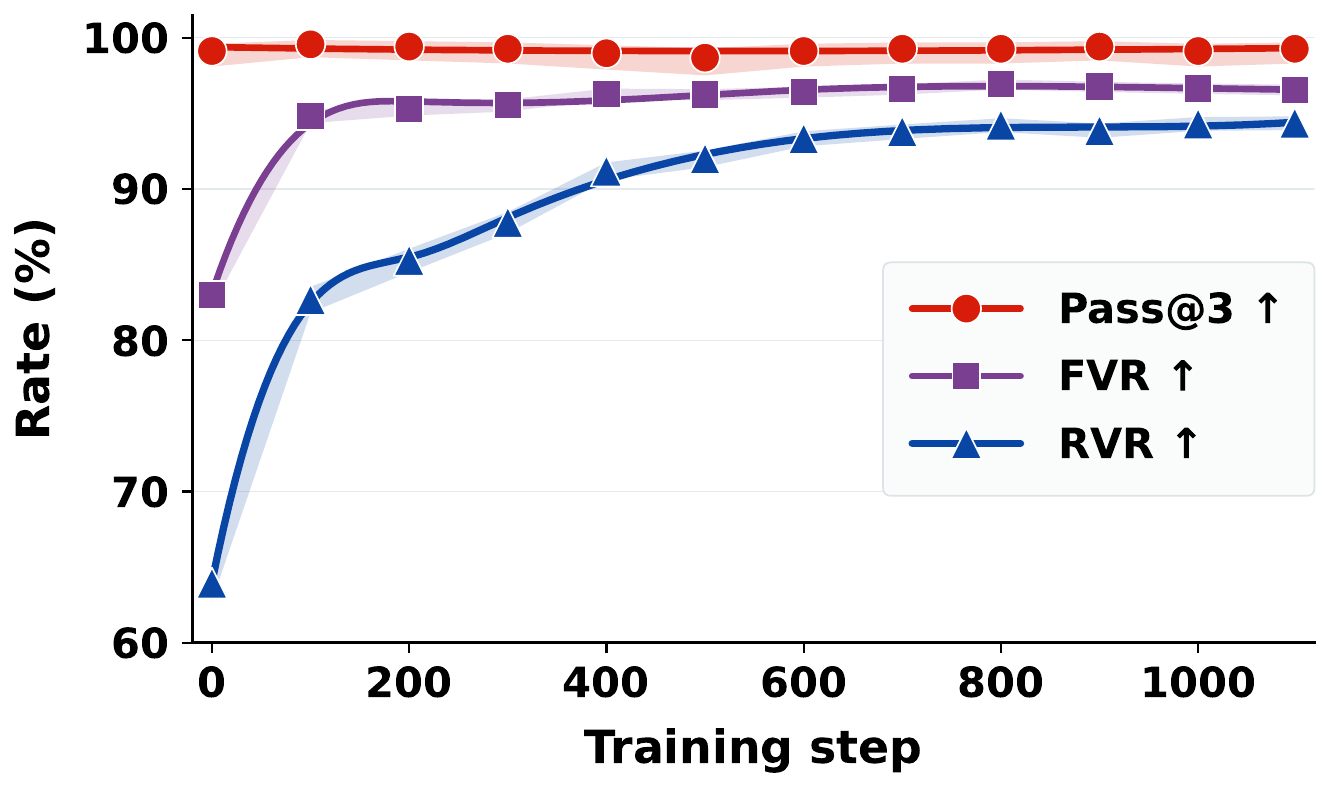}\\[-0.8ex]
        {\small\textbf{(a)} Performance during training}
    \end{minipage}\hfill
    \begin{minipage}[t]{0.32\linewidth}
        \centering
        \includegraphics[width=\linewidth]{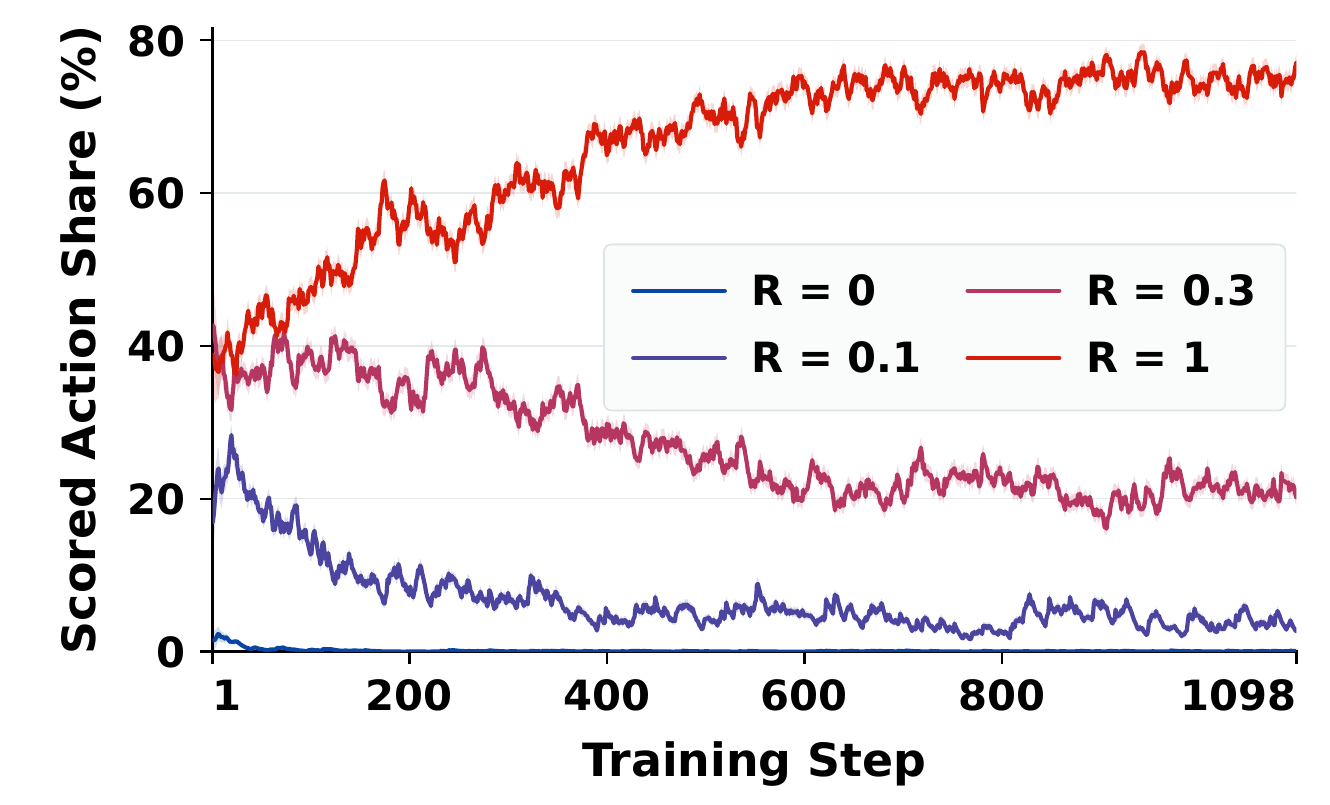}\\[-0.8ex]
        {\small\textbf{(b)} Verification-signal proportions}
    \end{minipage}\hfill
    \begin{minipage}[t]{0.32\linewidth}
        \centering
        \includegraphics[width=\linewidth]{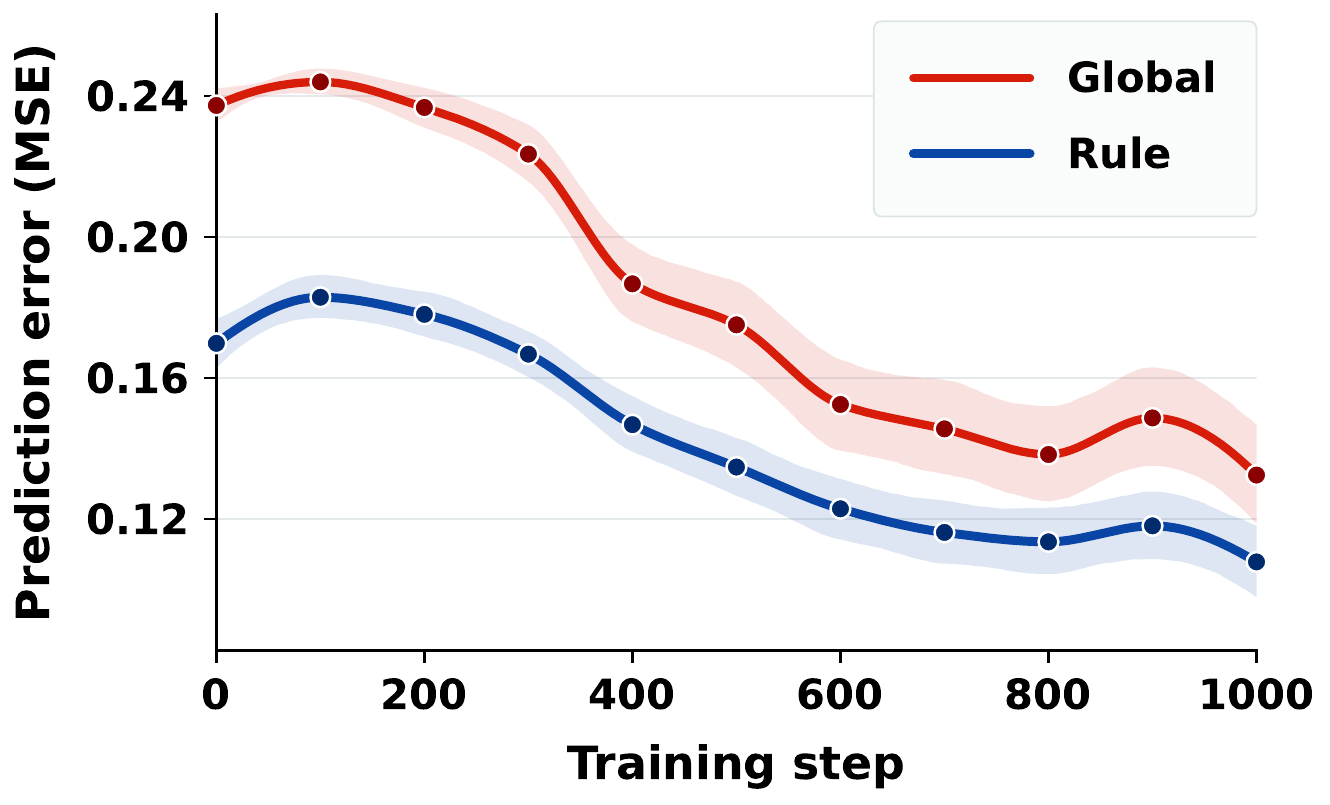}\\[-0.8ex]
        {\small\textbf{(c)} Centering-value error}
    \end{minipage}
    \setlength{\abovecaptionskip}{4pt}
    \caption{Training diagnostics of \method{}. Shaded regions indicate 95\% confidence intervals. (a) Performance during training, measured by Pass@3, FVR and RVR of Qwen3-8B on ProverQA. (b) Proportions of each verification signal of Qwen3-8B. (c) Prediction error of the global and rule-conditioned centering values during training.}
    \label{fig:training_diagnostics}
\end{figure}

\subsection{Discussion}
\label{sec:discussion}

\paragraph{Is \method{} effective across different backbone models?}
Table~\ref{tab:backbone_results} shows that \method{} improves answer
performance and rule-verified reasoning on Llama-3.1-8B and
GLM-Z1-9B-0414.
With Llama-3.1-8B, \method{} improves Avg@3 and Pass@3 over the backbone by
7.65 and 6.30 points, respectively.
With GLM-Z1-9B-0414, it improves FVR and RVR over SFT by 11.98 and 30.03
points, respectively, while reducing RGD by 0.23.
Across all four backbone models, \textit{\method{} outperforms both the backbone model and SFT on all five metrics on ProverQA}.

\paragraph{Does \method{} generalize to a broader inference system?}
ProverQA proof traces are constructed under the training-time rule system
$\mathcal{R}_{\mathrm{train}}$. The evaluated FOLIO subset contains valid
derivations that require a strictly broader system,
$\mathcal{R}_{\mathrm{FOLIO}}\supset\mathcal{R}_{\mathrm{train}}$, adding universal
and existential quantifier rule families absent from the training ontology
(Appendix~\ref{app:quantifier_coverage}).
Table~\ref{tab:cross_benchmark_results} reports the performance of
ProverQA-trained models on FOLIO and ProofWriter.
Notably, under $\mathcal{R}_{\mathrm{FOLIO}}$, ProverQA-trained \method{} achieves an Avg@3 of 90.85, an FVR of 82.58, and an RVR of 59.70 on FOLIO, compared with 88.89, 64.46, and 38.87 for the backbone model.
These results show that \method{} transfers beyond the
rule ontology used to construct ProverQA. \textit{The learned policy produces
answer-supporting proofs that remain verifiable under a broader inference system}.

\paragraph{How does \method{} improve response quality?}
Figure~\ref{fig:training_diagnostics}(a) shows the model's performance on
ProverQA throughout training.
Complementing this, Figure~\ref{fig:training_diagnostics}(b) shows how the distribution of verification signals evolves during training.
Answer accuracy remains high during training, while FVR increases to 96.56 and RVR increases to 94.36.
Meanwhile, the proportion of actions receiving the full verification signal
($R_{k,t}^{v}=1.0$) increases, whereas the proportions receiving $0.1$
or $0.3$ decline, indicating failures in the semantic, rule,
or proof-progress checks.
This observation is consistent with the action-level supervision provided
by \MCFV{}. Even among samples with correct answers, \textit{UNSAT-based
verification and \textsc{RuleCheck} distinguish the semantic validity (measured by FVR) and rule
faithfulness (measured by RVR) of individual proof steps, providing learning signals unavailable
from answer correctness alone}.

\paragraph{Can rule-conditioned centering values account for differences in verification difficulty?}
Reasoning actions in formal verification have different rule constraints
and dependency structures~\citep{saparov2023ood}. We investigate whether these structural properties characterize their verification difficulty?
Figure~\ref{fig:training_diagnostics}(c) compares a global-mean centering value (labeled Global)
with a rule-conditioned centering value (labeled Rule).
We cross-validate both centering-value estimators on data unseen during fitting.
The rule-conditioned centering value has lower prediction error than the global-mean centering value during training: at step 100, the error decreases from 0.24 to 0.18.
These results show that \textit{verification difficulty varies systematically with the logical operation performed by an action, rather than being uniform across actions}.
This provides empirical support for the rule-conditioned centering value used to center verification signals in \MCFV{}.

\Needspace{0.40\textheight}
\paragraph{How do \MCFV{} and \ASDC{} jointly shape the optimization direction?}\mbox{}\par
\begin{wrapfigure}{r}{0.49\linewidth}
    \vspace{-1\baselineskip}
    \centering
    \includegraphics[width=\linewidth]{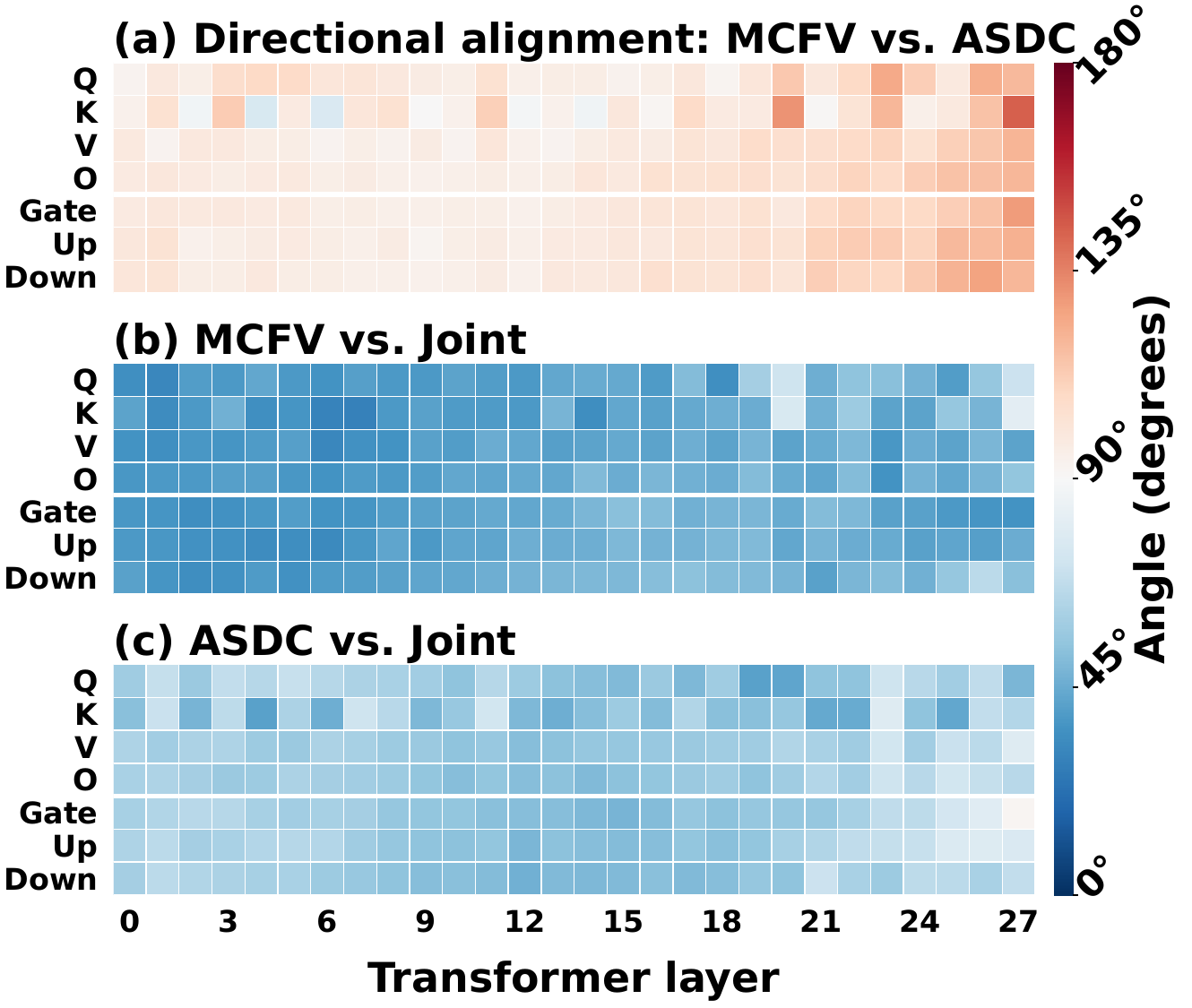}
    \setlength{\abovecaptionskip}{-10pt}
    \caption{Layer-wise and module-wise gradient at training on Qwen2.5-7B-Instruct.
    Figures show angles between (a) \MCFV{} and \ASDC{}, (b) \MCFV{} and the joint
    direction, and (c) \ASDC{} and the joint direction.
    }
    \label{fig:gradient_alignment}
    \vspace{-0.5\baselineskip}
\end{wrapfigure}

Figure~\ref{fig:gradient_alignment} illustrates the gradient relationships
between \MCFV{} and \ASDC{} across Transformer layers and parameter modules.
We compute the gradients of the two supervision signals and construct
the joint direction using their training weights.
Figure~\ref{fig:gradient_alignment}(a) shows that the two components are
nearly orthogonal in the early layers, with an angle of $96.6^\circ$
between the concatenated gradients of layers 0--8. Directional
competition increases with network depth, reaching
$126.6^\circ$ in the final layer.
This shows that the two supervision signals do not simply reinforce
the model along the same direction, but provide distinct optimization
signals in parameter space~\citep{yu2020gradient}.
Therefore, answer support and formal correctness are not equivalent
and require distinct supervision signals.
Figure~\ref{fig:gradient_alignment}(b)(c) further shows the weighted
joint direction.
Across the 196 layer--module parameter blocks, its angle with \MCFV{}
is acute in every block, and its angle with \ASDC{} is acute in 195 blocks.
The weighted \MCFV{} gradient norm is approximately 1.14 times that of
\ASDC{}, placing the two components at comparable magnitudes.
These results provide mechanistic evidence for verification-aligned optimization: although \MCFV{} and \ASDC{} induce distinct and sometimes competing gradients, \textit{verification-aligned optimization aligns their weighted contributions into a shared update direction that locally improves both objectives across almost all parameter blocks}.

\section{Limitations}
\label{sec:limitations}

Since our main claim concerns verifiable natural-language logical reasoning, evaluating \method{} on English benchmarks with well-defined deductive structures necessarily covers only a controlled subset of natural-language reasoning.
Such benchmarks necessarily abstract away some phenomena encountered in unrestricted natural language, including linguistic ambiguity, implicit background knowledge, and pragmatic interpretation. Moreover, while our evaluation covers multiple benchmarks, model families, and proof structures, it does not exhaust the broader reasoning landscape. Future work may investigate multilingual settings and additional logical formalisms.

\section{Conclusion}
\label{sec:conclusion}

We introduce \method{}, a formal verification-driven reinforcement learning framework for verifiable natural-language logical reasoning.
The framework combines UNSAT-based \textit{machine-checkable formal verification} with
an \textit{answer-supporting dependency closure} to construct verified proof states and
align policy learning with the final derivation.
Experiments across multiple backbone models and benchmarks show that \method{} improves answer performance and verification pass rates for reasoning steps.
This work integrates formal verification into proof construction and policy optimization on large language models, enabling models to perform verifiable proof-based reasoning over natural-language logical problems.

\section*{AI Use Statement}
In the preparation of this paper, GPT5.6-Sol was used exclusively for the purpose of polishing and refining the linguistic style of certain sentences to improve readability and fluency in section~\ref{sec:related_work} and~\ref{sec:limitations} and appendix~\ref{app:cross_solver_validation} and~\ref{app:implementation}.
As described in Appendix~\ref{app:cross_solver_validation}, GPT-5.5 was used as an independent semantic reviewer to cross-check solver judgments.
All core intellectual contributions are the product
of the authors’ own work, including but not limited to:

\begin{itemize}[leftmargin=1.5em,topsep=-1pt,partopsep=0pt]
    \item Original ideas and novel viewpoints. Any innovative
concepts, hypotheses, and critical analysis are the product
of the authors’ own work.
    \item Experimental results. All code, figures, findings,
and interpretations thereof were generated, designed,
collected, and analyzed by the authors.
    \item Substantive descriptions of the work. The description of
the methodology, experimental procedures, and discussion
of the research implications were entirely authored by the
research team.
\end{itemize}

\section*{Reproducibility Statement}
The datasets, backbone models, and baselines are described
in Section~\ref{sec:experiments}, with metric definitions provided in
Appendix~\ref{app:evaluation_metrics}.  The complete policy objective,
verification specifications, inference systems, solver-validation procedures,
implementation details, training configuration, computational cost, and prompt
templates are documented in Appendices~\ref{app:policy_implementation}--\ref{app:prompt}.
Source code is available at \url{https://anonymous.4open.science/r/Proof-R1/}.

\bibliography{iclr2027_conference}
\bibliographystyle{iclr2027_conference}

\appendix
\startcontents[appendices]

\clearpage
\begingroup
\definecolor{appendixcontentsblue}{RGB}{42,91,176}
\colorlet{appendixcontentsred}{appendixcontentsblue}%
\thispagestyle{plain}
\hypersetup{hidelinks}

\vspace*{-1.4em}
{\fontsize{16}{19.2}\selectfont\bfseries Appendix\par}

\vspace{1.8em}
{\fontsize{11}{13.2}\selectfont\bfseries Contents of Appendix\par}
\vspace{0.25em}
\hrule height 0.5pt
\vspace{0.55em}

\titlecontents{section}
  [3.3em]
  {\addvspace{0.20em}\color{appendixcontentsblue}\fontsize{10.5}{11.5}\selectfont\bfseries}
  {\contentslabel{2.6em}}
  {}
  {\color{black}\hfill\bfseries\contentspage}

\titlecontents{subsection}
  [5.2em]
  {\addvspace{0em}\color{appendixcontentsblue}\fontsize{10}{10.5}\selectfont\normalfont}
  {\contentslabel{3.7em}}
  {}
  {\color{black}\titlerule*[0.65pc]{.}\contentspage}

\printcontents[appendices]{}{1}{\setcounter{tocdepth}{2}}

\vspace{0.45em}
\hrule height 0.5pt
\endgroup

\section{Policy Optimization Implementation}
\label{app:policy_implementation}

Let $L_k$ and $T_k$ denote the response length and number of structured
actions in rollout $k$, respectively. The normalized action-to-token alignment
matrix used in Equation~\ref{eq:advantage_projection} is
\begin{equation}
M^k_{j,t}=
\frac{\ind[j\in\operatorname{Span}(a_t^k)]}
{|\operatorname{Span}(a_t^k)|},
\qquad
M^k\in\mathbb{R}^{L_k\times T_k}.
\label{eq:action_token_matrix}
\end{equation}
Thus, its $t$-th column distributes the advantage of action $a_t^k$
uniformly over the token positions in $\operatorname{Span}(a_t^k)$ and
assigns zero weight elsewhere.
For token $w_j^k$ with context $h_j^k=(x^k,w_{<j}^k)$, the policy ratio is
$r_{k,j}(\theta)=\pi_\theta(w_j^k\mid h_j^k)\big/
\pi_{\theta_{\mathrm{old}}}(w_j^k\mid h_j^k)$, where $\mathcal{B}$ indexes
rollouts sampled from the old policy $\pi_{\theta_{\mathrm{old}}}$.
The optimized objective is the KL-regularized clipped loss
\begin{align}
\mathcal{L}_{\mathrm{clip}}(\theta)
={}&-\frac{1}{N_{\mathcal{B}}}
\sum_{k\in\mathcal{B}}\sum_{j=1}^{L_k}
\min\Bigl\{
r_{k,j}(\theta)A_j^{\mathrm{tok},k},\nonumber\\
&\hspace{30mm}
\operatorname{clip}\!\left(r_{k,j}(\theta),
1-\epsilon_-,1+\epsilon_+\right)A_j^{\mathrm{tok},k}
\Bigr\}\nonumber\\
&+\frac{\lambda_{\mathrm{KL}}}{|\mathcal{B}|}
\sum_{k\in\mathcal{B}}\sum_{j=1}^{L_k}d_{k,j}(\theta).
\label{eq:policy_objective}
\end{align}
Here $\epsilon_-$ and $\epsilon_+$ are the clipping thresholds,
$N_{\mathcal{B}}$ is the number of mapped actions, clamped to at least one,
and $\lambda_{\mathrm{KL}}$ controls reference-policy regularization.
The sampled KL penalty is
\begin{equation}
d_{k,j}(\theta)
=\frac{\pi_{\mathrm{ref}}(w_j^k\mid h_j^k)}
       {\pi_\theta(w_j^k\mid h_j^k)}
-\log\frac{\pi_{\mathrm{ref}}(w_j^k\mid h_j^k)}
          {\pi_\theta(w_j^k\mid h_j^k)}-1,
\label{eq:sampled_kl}
\end{equation}
where $\pi_{\mathrm{ref}}$ is the fixed reference policy.

Let $g(r,A)=\min\{rA,\operatorname{clip}(r,1-\epsilon_-,1+\epsilon_+)A\}$
denote the surrogate term in Equation~\ref{eq:policy_objective}.
The current optimizer additionally applies dual clipping to negative
advantages, replacing $g$ with
\begin{equation}
g_{\mathrm{dual}}(r,A)=
\begin{cases}
\max\{g(r,A),c_{\mathrm{dual}}A\},&A<0,\\
g(r,A),&A\geq 0,
\end{cases}
\qquad c_{\mathrm{dual}}>1.
\label{eq:dual_clip}
\end{equation}
The implemented policy loss is therefore
\begin{equation}
\mathcal{L}_{\mathrm{impl}}(\theta)
=-\frac{1}{N_{\mathcal{B}}}\sum_{k\in\mathcal{B}}\sum_{j=1}^{L_k}
g_{\mathrm{dual}}\!\left(r_{k,j}(\theta),A_j^{\mathrm{tok},k}\right)
+\frac{\lambda_{\mathrm{KL}}}{|\mathcal{B}|}
\sum_{k\in\mathcal{B}}\sum_{j=1}^{L_k}d_{k,j}(\theta).
\label{eq:implemented_policy_loss}
\end{equation}
Group-level and negative-mass controls are applied to action credit before
the token projection.
VERL sums token losses within each response and averages over responses,
multiplying the projected advantages by $|\mathcal{B}|/N_{\mathcal{B}}$
therefore yields the action-normalized policy term above.

\section{Evaluation Metric Definitions}
\label{app:evaluation_metrics}

Let $\mathcal{D}_{\mathrm{eval}}$ contain $N$ evaluation problems, let $K$
denote the number of sampled outputs per problem, and let
$r_{i,k}\in\{0,1\}$ indicate whether response $k$ for problem $i$ gives the
correct answer.
Avg@K and Pass@K for answer correctness is measured by
\begin{align}
    \mathrm{Avg@}K
    &= \frac{1}{NK}\sum_{i=1}^{N}\sum_{k=1}^{K}r_{i,k},
    \label{eq:avg_at_k}\\
    \mathrm{Pass@}K
    &= \frac{1}{N}\sum_{i=1}^{N}
    \mathbf{1}\!\left[\sum_{k=1}^{K}r_{i,k}>0\right].
    \label{eq:pass_at_k}
\end{align}
We use $K=3$ in all reported experiments.

For proof verification, let $\mathcal{E}_{\mathrm{sem}}$ and
$\mathcal{E}_{\mathrm{rule}}$ denote the generated proof steps eligible for
semantic and rule verification at the corresponding verification level.
For each eligible step $e$, the Boolean verdicts
$v_e^{\mathrm{sem}},v_e^{\mathrm{rule}}\in
\{\mathrm{true},\mathrm{false}\}$ record, respectively, whether the step is
semantically valid and whether it is faithful to its declared rule.
The two micro-averaged step pass rates are:
\begin{align}
    \mathrm{FVR}
    &= \frac{\sum_{e\in\mathcal{E}_{\mathrm{sem}}}
    \mathbf{1}\!\left[v_e^{\mathrm{sem}}\right]}
    {|\mathcal{E}_{\mathrm{sem}}|},
    \label{eq:fvr_metric}\\
    \mathrm{RVR}
    &= \frac{\sum_{e\in\mathcal{E}_{\mathrm{rule}}}
    \mathbf{1}\!\left[v_e^{\mathrm{rule}}\right]}
    {|\mathcal{E}_{\mathrm{rule}}|}.
    \label{eq:rvr_metric}
\end{align}
When an eligible set is empty, its rate is defined as zero.

For an output with an available reference proof, let
$N_{i,k}^{\mathrm{generated}}$ be the number of generated intermediate
conclusions in its answer-supporting dependency closure and let $N_i^{\mathrm{reference}}$ be the number of reference proof steps.
Its granularity deviation and the dataset-level RGD are:
\begin{align}
    \mathrm{RGD}_{i,k}
    &= \left|\log\frac{\max(1,N_{i,k}^{\mathrm{generated}})}
    {\max(1,N_i^{\mathrm{reference}})}\right|,
    \label{eq:rgd_output}\\
    \mathrm{RGD}
    &= \frac{1}{|\mathcal{Q}|}
    \sum_{(i,k)\in\mathcal{Q}}\mathrm{RGD}_{i,k},
    \label{eq:rgd}
\end{align}
where $\mathcal{Q}$ is the set of evaluated outputs with an available reference proof.

\section{Additional ProverQA Results}
\label{app:additional_results}

\subsection{Easy and Medium Difficulty Splits of Main Results}

\begin{table}[!ht]
    \setlength{\belowcaptionskip}{2.5pt}
    \caption{Additional results of Qwen2.5-7B-Instruct and Qwen3-8B on the ProverQA Easy and Medium splits, reported as means followed by standard deviations in gray. \captiontrainfreeicon{} and \captiontrainingicon{} denote training-free and training-based methods, respectively. \textbf{Bold} and \underline{underline} denote the best and second-best means within each backbone model and split.}
    \label{tab:additional_difficulty_results}
    \centering
    \fontsize{8}{8.5}\selectfont
    \setlength{\tabcolsep}{1pt}
    \renewcommand{\arraystretch}{1.2}
    \resizebox{\linewidth}{!}{%
    \begin{tabular}{@{}cl*{5}{c}@{\hspace{5pt}}*{5}{c}@{}}
        \toprule
        \multicolumn{1}{c}{\multirow{2}{*}{\textbf{Model}}} & \multicolumn{1}{c}{\multirow{2}{*}{\textbf{Method}}} & \multicolumn{5}{c}{\textbf{ProverQA-Easy}}
        & \multicolumn{5}{c}{\textbf{ProverQA-Medium}} \\
        \cmidrule(lr){3-7}\cmidrule(lr){8-12}
        & & Avg@3 $\uparrow$ & Pass@3 $\uparrow$ & FVR $\uparrow$ & RVR $\uparrow$ & RGD $\downarrow$
        & Avg@3 $\uparrow$ & Pass@3 $\uparrow$ & FVR $\uparrow$ & RVR $\uparrow$ & RGD $\downarrow$ \\
        \midrule
        \multirow{7}{*}[-2pt]{\rotatebox[origin=c]{90}{\tablemodelname{Qwen2.5-7B-Instruct}}}
        & \trainfreeicon Backbone & \meanstdcell{62.20}{1.33} & \meanstdcell{87.36}{2.06} & \meanstdcell{77.37}{0.66} & \meanstdcell{31.19}{2.17} & \meanstdcell{0.51}{0.016} & \meanstdcell{40.12}{0.99} & \meanstdcell{61.09}{3.27} & \meanstdcell{\underline{64.82}}{0.71} & \meanstdcell{14.45}{1.10} & \meanstdcell{0.90}{0.013} \\
        & \trainfreeicon LogicAgent & \meanstdcell{76.50}{0.68} & \meanstdcell{91.57}{1.72} & \meanstdcell{68.30}{1.83} & \meanstdcell{\underline{46.17}}{1.23} & \meanstdcell{\underline{0.46}}{0.010} & \meanstdcell{56.26}{0.84} & \meanstdcell{82.81}{2.54} & \meanstdcell{51.32}{1.63} & \meanstdcell{\underline{25.57}}{1.37} & \meanstdcell{0.97}{0.011} \\
        & \trainfreeicon GoV & \meanstdcell{69.09}{1.51} & \meanstdcell{90.42}{1.82} & \meanstdcell{76.67}{1.06} & \meanstdcell{37.46}{1.26} & \meanstdcell{0.47}{0.016} & \meanstdcell{39.82}{0.91} & \meanstdcell{64.25}{3.20} & \meanstdcell{61.27}{0.90} & \meanstdcell{19.39}{0.53} & \meanstdcell{0.92}{0.019} \\
        \addlinespace[2pt]
        & \trainingicon SFT & \meanstdcell{64.88}{2.11} & \meanstdcell{89.27}{1.91} & \meanstdcell{76.58}{0.71} & \meanstdcell{34.45}{2.18} & \meanstdcell{0.51}{0.012} & \meanstdcell{40.57}{1.23} & \meanstdcell{64.25}{3.22} & \meanstdcell{63.33}{1.01} & \meanstdcell{14.29}{1.22} & \meanstdcell{0.82}{0.024} \\
        \addlinespace[2pt]
        & \trainingicon GRPO & \meanstdcell{66.54}{1.90} & \meanstdcell{92.72}{1.61} & \meanstdcell{\underline{83.60}}{0.50} & \meanstdcell{21.42}{1.45} & \meanstdcell{0.54}{0.007} & \meanstdcell{46.00}{1.57} & \meanstdcell{81.00}{2.65} & \meanstdcell{62.79}{1.58} & \meanstdcell{8.94}{0.90} & \meanstdcell{1.03}{0.009} \\
        & \trainingicon PRoSFI & \meanstdcell{\underline{77.08}}{1.22} & \meanstdcell{\underline{96.36}}{0.00} & \meanstdcell{77.65}{0.17} & \meanstdcell{23.03}{2.31} & \meanstdcell{0.55}{0.019} & \meanstdcell{\underline{57.77}}{0.15} & \meanstdcell{\textbf{86.65}}{1.80} & \meanstdcell{58.09}{0.27} & \meanstdcell{14.86}{1.06} & \meanstdcell{\underline{0.79}}{0.010} \\
        & \trainingicon\textbf{\method{}} & \meanstdcell{\textbf{79.31}}{0.80} & \meanstdcell{\textbf{96.55}}{1.13} & \meanstdcell{\textbf{84.35}}{0.55} & \meanstdcell{\textbf{68.67}}{1.08} & \meanstdcell{\textbf{0.40}}{0.016} & \meanstdcell{\textbf{59.43}}{0.66} & \meanstdcell{\underline{85.07}}{2.39} & \meanstdcell{\textbf{68.30}}{1.07} & \meanstdcell{\textbf{34.13}}{0.52} & \meanstdcell{\textbf{0.73}}{0.030} \\
        \midrule
        \multirow{7}{*}[-2pt]{\rotatebox[origin=c]{90}{\tablemodelname{Qwen3-8B}}}
        & \trainfreeicon Backbone & \meanstdcell{97.06}{0.13} & \meanstdcell{\textbf{100.00}}{0.00} & \meanstdcell{82.35}{0.32} & \meanstdcell{69.46}{0.12} & \meanstdcell{0.30}{0.012} & \meanstdcell{96.83}{0.26} & \meanstdcell{\underline{99.55}}{0.45} & \meanstdcell{82.16}{1.50} & \meanstdcell{59.88}{0.92} & \meanstdcell{0.31}{0.011} \\
        & \trainfreeicon LogicAgent & \meanstdcell{\textbf{99.36}}{0.13} & \meanstdcell{\underline{99.62}}{0.38} & \meanstdcell{97.08}{0.43} & \meanstdcell{90.35}{0.43} & \meanstdcell{0.27}{0.004} & \meanstdcell{\textbf{98.94}}{0.40} & \meanstdcell{\textbf{100.00}}{0.00} & \meanstdcell{\underline{95.02}}{0.36} & \meanstdcell{81.38}{0.55} & \meanstdcell{0.29}{0.012} \\
        & \trainfreeicon GoV & \meanstdcell{98.08}{0.44} & \meanstdcell{99.23}{0.54} & \meanstdcell{93.12}{0.93} & \meanstdcell{81.43}{0.63} & \meanstdcell{0.26}{0.002} & \meanstdcell{97.59}{0.15} & \meanstdcell{98.64}{0.78} & \meanstdcell{89.32}{0.61} & \meanstdcell{70.43}{1.23} & \meanstdcell{0.26}{0.003} \\
        \addlinespace[2pt]
        & \trainingicon SFT & \meanstdcell{96.42}{0.51} & \meanstdcell{\underline{99.62}}{0.38} & \meanstdcell{87.12}{1.43} & \meanstdcell{74.58}{0.86} & \meanstdcell{0.30}{0.006} & \meanstdcell{\underline{97.74}}{0.69} & \meanstdcell{\underline{99.55}}{0.45} & \meanstdcell{83.88}{0.92} & \meanstdcell{62.37}{0.45} & \meanstdcell{0.31}{0.004} \\
        \addlinespace[2pt]
        & \trainingicon GRPO & \meanstdcell{\underline{99.23}}{0.22} & \meanstdcell{\textbf{100.00}}{0.00} & \meanstdcell{98.27}{0.05} & \meanstdcell{92.17}{0.11} & \meanstdcell{0.28}{0.001} & \meanstdcell{97.59}{0.15} & \meanstdcell{\underline{99.55}}{0.45} & \meanstdcell{94.74}{0.52} & \meanstdcell{69.31}{1.13} & \meanstdcell{0.26}{0.003} \\
        & \trainingicon PRoSFI & \meanstdcell{98.72}{0.00} & \meanstdcell{\underline{99.62}}{0.38} & \meanstdcell{\underline{98.47}}{0.16} & \meanstdcell{\underline{93.49}}{0.15} & \meanstdcell{\underline{0.26}}{0.005} & \meanstdcell{96.98}{0.73} & \meanstdcell{98.64}{0.45} & \meanstdcell{93.21}{0.31} & \meanstdcell{\underline{88.24}}{0.45} & \meanstdcell{\underline{0.26}}{0.004} \\
        & \trainingicon\textbf{\method{}} & \meanstdcell{98.72}{0.34} & \meanstdcell{\textbf{100.00}}{0.00} & \meanstdcell{\textbf{99.19}}{0.18} & \meanstdcell{\textbf{97.46}}{0.30} & \meanstdcell{\textbf{0.25}}{0.012} & \meanstdcell{97.44}{0.15} & \meanstdcell{98.64}{0.78} & \meanstdcell{\textbf{98.07}}{0.04} & \meanstdcell{\textbf{96.61}}{0.14} & \meanstdcell{\textbf{0.25}}{0.005} \\
        \bottomrule
    \end{tabular}}
\end{table}

Table~\ref{tab:additional_difficulty_results} complements the results with the Easy and Medium splits of
ProverQA. Results are averaged over 3 independent seeds.

On Qwen2.5-7B-Instruct, \method{} achieves the best Avg@3, FVR, RVR,
and RGD on both difficulty splits, and also obtains the best Pass@3 on
Easy.  The largest separation appears in rule-grounded verification:
\method{} improves RVR over the strongest baseline by 22.50 points on Easy
and 8.56 points on Medium.  These results show that the verification gains in
the main table persist across easier reasoning regimes rather than arising
only from the aggregate or Hard split.

On Qwen3-8B, answer accuracy is already near saturation across methods, while
\method{} consistently provides the strongest proof-verification quality.
It achieves the best FVR, RVR, and RGD on both Easy and Medium. Relative to
the strongest baseline, its RVR improves by 3.97 points on Easy and 8.37
points on Medium. The same pattern across both backbone models shows that
\method{} improves the validity and rule fidelity of generated proofs throughout the difficulty range.

\subsection{Cross-Dataset Results with Qwen2.5-7B-Instruct}
\label{app:qwen25_cross_dataset}

The FOLIO evaluation uses the broader trusted inference system
$\mathcal{R}_{\mathrm{FOLIO}}\supset\mathcal{R}_{\mathrm{train}}$ described in
Appendix~\ref{app:quantifier_coverage}. Table~\ref{tab:qwen25_cross_dataset}
reports the performance of ProverQA-trained Qwen2.5-7B-Instruct models on
FOLIO and ProofWriter.

On FOLIO, \method{} achieves the best Avg@3 and RVR, improving over the
backbone model by 13.08 and 13.43 points, respectively, while attaining the
joint-best Pass@3 of 84.31. On ProofWriter, it improves Avg@3, Pass@3, FVR,
and RVR over the backbone model by 18.66, 21.34, 6.54, and 15.72 points,
respectively. Together with the Qwen3-8B results in
Table~\ref{tab:cross_benchmark_results}, these results show that \emph{the
cross-dataset gains of ProverQA-trained \method{} persist across two backbone models}.

\begin{table}[!ht]
    \setlength{\belowcaptionskip}{2.5pt}
    \caption{Additional cross-dataset results of ProverQA-trained Qwen2.5-7B-Instruct models on FOLIO and ProofWriter. \captiontrainfreeicon{} and \captiontrainingicon{} denote training-free and training-based methods, respectively. \textbf{Bold} and \underline{underline} denote the best and second-best results within each dataset.}
    \label{tab:qwen25_cross_dataset}
    \centering
    \fontsize{8}{8.5}\selectfont
    \setlength{\tabcolsep}{1.5pt}
    \renewcommand{\arraystretch}{1.2}
    \begin{tabular*}{\linewidth}{@{\extracolsep{\fill}}c@{\hspace{4pt}}l*{4}{c}@{}}
        \toprule
        \multicolumn{1}{c}{\textbf{Dataset}} & \multicolumn{1}{c}{\textbf{Method}} & Avg@3 $\uparrow$ & Pass@3 $\uparrow$ & FVR $\uparrow$ & RVR $\uparrow$ \\
        \midrule
        \multirow{3}{*}[-1pt]{\rotatebox[origin=c]{90}{\tablemodelname{FOLIO}}}
        & \trainfreeicon Backbone & 49.67 & 72.55 & \underline{68.97} & 15.04 \\
        & \trainingicon SFT & \underline{58.17} & \textbf{84.31} & 68.33 & \underline{18.78} \\
        & \trainingicon\textbf{\method{}} & \textbf{62.75} & \textbf{84.31} & \textbf{69.30} & \textbf{28.47} \\
        \midrule
        \multirow{3}{*}[-1pt]{\rotatebox[origin=c]{90}{\tablemodelname{\shortstack{Proof\\Writer}}}}
        & \trainfreeicon Backbone & 48.67 & \underline{69.33} & \underline{66.42} & \underline{15.55} \\
        & \trainingicon SFT & \underline{50.22} & 68.67 & 65.15 & 14.91 \\
        & \trainingicon\textbf{\method{}} & \textbf{67.33} & \textbf{90.67} & \textbf{72.96} & \textbf{31.27} \\
        \bottomrule
    \end{tabular*}
\end{table}

\subsection{Hard-Split Results across Additional Backbone Models}
\label{app:additional_hard_backbones}

\begin{table}[!ht]
    \setlength{\belowcaptionskip}{2.5pt}
    \caption{Additional ProverQA-Hard results across backbone models. \captiontrainfreeicon{} and \captiontrainingicon{} denote training-free and training-based methods, respectively. \textbf{Bold} and \underline{underline} denote the best and second-best results within each backbone model.}
    \label{tab:additional_hard_backbones}
    \centering
    \fontsize{8}{8.5}\selectfont
    \setlength{\tabcolsep}{1pt}
    \renewcommand{\arraystretch}{1.2}
    \begin{tabular*}{\linewidth}{@{\extracolsep{\fill}}c@{\hspace{4pt}}l*{5}{c}@{}}
        \toprule
        \multicolumn{1}{c}{\textbf{Model}} & \multicolumn{1}{c}{\textbf{Method}} & Avg@3 $\uparrow$ & Pass@3 $\uparrow$ & FVR $\uparrow$ & RVR $\uparrow$ & RGD $\downarrow$ \\
        \midrule
        \multirow{3}{*}[-1pt]{\rotatebox[origin=c]{90}{\tablemodelname{\shortstack{Llama-3.1\\8B}}}}
        & \trainfreeicon Backbone & \underline{35.94} & \underline{68.33} & \underline{42.21} & 5.74 & \underline{0.85} \\
        & \trainingicon SFT & 32.03 & 63.70 & 36.18 & \underline{6.66} & \textbf{0.83} \\
        & \trainingicon\textbf{\method{}} & \textbf{38.79} & \textbf{71.89} & \textbf{43.29} & \textbf{11.06} & 0.87 \\
        \midrule
        \multirow{3}{*}[-1pt]{\rotatebox[origin=c]{90}{\tablemodelname{\shortstack{GLM-Z1\\9B-0414}}}}
        & \trainfreeicon Backbone & 62.31 & 84.92 & \underline{76.00} & 41.89 & 1.01 \\
        & \trainingicon SFT & \underline{67.34} & \underline{94.97} & 75.30 & \underline{45.21} & \underline{0.90} \\
        & \trainingicon\textbf{\method{}} & \textbf{80.23} & \textbf{98.49} & \textbf{90.79} & \textbf{83.24} & \textbf{0.65} \\
        \bottomrule
    \end{tabular*}
\end{table}

Table~\ref{tab:additional_hard_backbones} extends the ProverQA-Hard
comparison to Llama-3.1-8B and GLM-Z1-9B-0414.  Across both backbone
models, \method{} achieves the best Avg@3, Pass@3, FVR, and RVR.  Its RVR
exceeds the strongest baseline by 4.40 points on Llama-3.1-8B and 38.03
points on GLM-Z1-9B-0414. On GLM-Z1-9B-0414, it also obtains the lowest
RGD.  These results extend \emph{the answer and verification gains on the most
challenging split beyond the Qwen model family}.

\section{Conditional Soundness of the Verified Proof Certificate}
\label{app:soundness}

\paragraph{Guarantee.}
Let $\Pi_\tau=G_\tau[\mathcal{C}_\tau(c_g)]$ be the answer-supporting proof
certificate defined in Equation~\ref{eq:proof_certificate}.  Assume faithful
formal representations, consistent premises, acyclic dependencies that resolve only to original premises or earlier actions, and trusted parsing, solver, and rule-checking components. If every generated conclusion in $\Pi_\tau$ discharges its semantic proof obligation, then:
\begin{equation}
\bigwedge_{c_t\in\mathcal{C}_\tau(c_g)\setminus\mathcal{P}}
v_t^{\mathrm{sem}}
\quad\Longrightarrow\quad
\mathcal{P}\models c_g.
\label{eq:conditional_semantic_guarantee}
\end{equation}
A valid \texttt{GOAL\_BINDING} identifies $c_g$ with the selected proposition
$\tilde y$, yielding $\mathcal{P}\models\tilde y$.  If every action in the
certificate also discharges its rule obligation, each explicit inference
additionally conforms to its declared rule schema.

\paragraph{Proof.}
Order the generated conclusions in the closure topologically as
$c_{t_1},\ldots,c_{t_m}=c_g$, and let $a_{t_i}$ denote the action that
generates $c_{t_i}$.  By assumption, every dependency resolves to an original
premise or an earlier action in this order.
For the base case, $c_{t_1}$ has no generated-conclusion ancestors, so the
resolved dependencies of $a_{t_1}$ are original premises.  Its discharged
semantic obligation (Equation~\ref{eq:semantic_verdict}) establishes that
these dependencies entail $c_{t_1}$, and hence
$\mathcal{P}\models c_{t_1}$.
For the induction step, suppose that $\mathcal{P}$ entails all earlier
generated conclusions in the closure.  Every dependency of $a_{t_i}$ is then
entailed by $\mathcal{P}$, while the discharged obligation establishes that
those dependencies entail $c_{t_i}$.  Transitivity gives
$\mathcal{P}\models c_{t_i}$.  Applying the same argument through
$c_{t_m}=c_g$ proves Equation~\ref{eq:conditional_semantic_guarantee}, and
valid goal binding gives $\mathcal{P}\models\tilde y$.

\section{Cross-Backend Verification and Independent Semantic Review}
\label{app:cross_solver_validation}

\paragraph{Evaluation protocol.}
The verification gains of \method{} remain unchanged when Z3 is replaced
with cvc5~\citep{barbosa2022cvc5},
veriT,\footnote{\url{https://www.verit-solver.org/}} or Isabelle/HOL\footnote{\url{https://isabelle.in.tum.de/doc/nitpick.pdf}}.
We recheck the frozen Qwen2.5-7B-Instruct outputs from the main ProverQA
comparison: Backbone, SFT, and \method{} each contribute three responses to
the same 681 problems, totaling 6,129 responses.
We use Z3 (\textit{versions 4.16.0}), cvc5 (\textit{versions 1.4.0}), and veriT (\textit{versions 2021.06.2-rmx}),
with a 5-second timeout per query. cvc5 uses logic \texttt{ALL} and
\texttt{finite-model-find=true}; veriT uses \texttt{UF} for quantified
inputs and \texttt{QF\_UF} otherwise.
Isabelle2025-2/HOL uses \texttt{blast} (at most 1 second), followed,
if needed, by \texttt{Nitpick} over domain sizes 1--10,
within a shared 5-second budget per query.

\begin{figure}[H]
    \centering
    \includegraphics[width=0.9\linewidth]{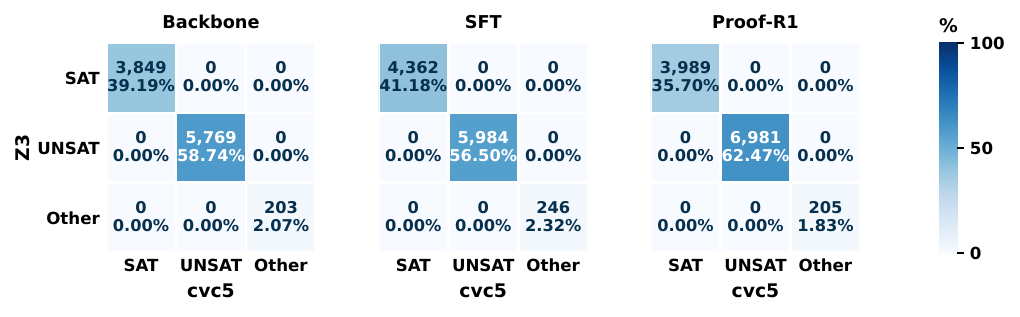}
    \caption{Z3--cvc5 paired local-check outcomes for Qwen2.5-7B-Instruct 
    outputs from Backbone, SFT, and \method{}. Rows denote Z3 and columns denote cvc5, using
    \texttt{SAT}/\texttt{UNSAT}/\texttt{Other}. Each cell shows its count
    and percentage of that method's local check attempts.}
    \label{fig:cross_solver_cvc5}
\end{figure}

\begin{figure}[H]
    \centering
    \includegraphics[width=0.9\linewidth]{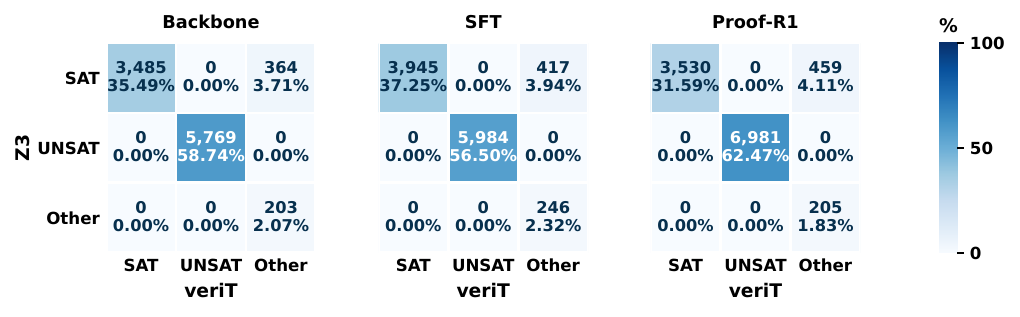}
    \caption{Z3--veriT paired local-check outcomes for the same frozen
    Qwen2.5-7B-Instruct outputs. Rows denote Z3 and columns denote veriT,
    using \texttt{SAT}/\texttt{UNSAT}/\texttt{Other}. Each cell shows its
    count and percentage of that method's local check attempts.}
    \label{fig:cross_solver_verit}
\end{figure}

\paragraph{Paired local checks.}
The three SMT backends check the same semantic verification condition from
Equation~\ref{eq:semantic_verdict}.
For this comparison, dependencies resolve against the original premises and
the same raw earlier conclusions.
We record \texttt{SAT}, \texttt{UNSAT}, and \texttt{Other} separately.
Of the 6,129 responses, 5,844 contain a parseable, nonempty structured
summary, yielding 31,588 local check attempts.
Z3 and cvc5 agree on all 30,934 determinate outcomes: 12,200 \texttt{SAT}
and 18,734 \texttt{UNSAT} (Figure~\ref{fig:cross_solver_cvc5}).
veriT confirms all 18,734 \texttt{UNSAT} outcomes and 10,960 \texttt{SAT}
outcomes, with no contradictory decision among the 29,694 jointly
determinate pairs (Figure~\ref{fig:cross_solver_verit}).
The other 1,240 Z3-\texttt{SAT} queries contain quantifiers and return
\texttt{unknown} in veriT; they are retained as \texttt{Other}.
All pipelines additionally share 654 input-processing failures:
525 formula-parsing failures, 128 missing dependencies, and one duplicate
identifier. No SMT solver timeout occurs in this fixed-query audit.

\paragraph{Native proofs and countermodels.}
Isabelle2025-2/HOL validates the same local queries through native
\texttt{blast} proof search and Nitpick countermodel search, without
calling an SMT backend.
A separate encoder translates the shared FOL syntax tree directly to HOL,
preserving quantifier scope, free constants, and nonempty-domain semantics.
An oracle-free kernel proof of the requested entailment establishes
\texttt{UNSAT}; it must have no open hypotheses or remaining subgoals.
Nitpick uses Kodkod/SAT4J over domain sizes 1--10, and establishes
\texttt{SAT} only when its final outcome is \texttt{genuine} and a
countermodel is saved. These model-search results are distinct from
kernel-certified theorems; an unsuccessful finite search remains
\texttt{Other}.
Each distinct query has a 5-second total budget, with at most 1 second
for \texttt{blast} and the remaining time for Nitpick.

\begin{figure}[H]
    \centering
    \includegraphics[width=0.9\linewidth]{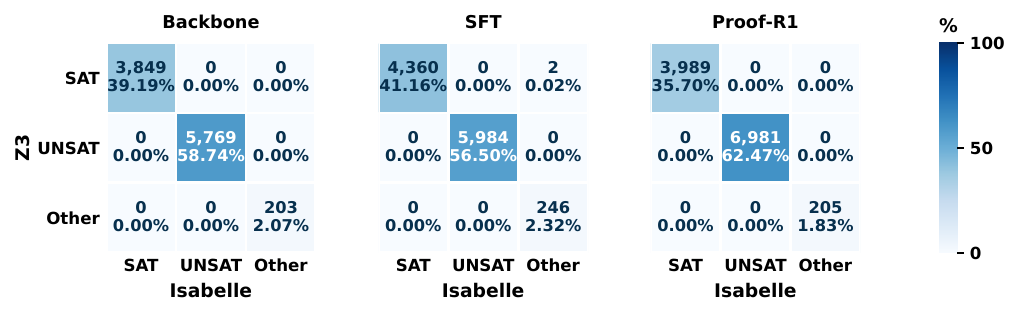}
    \caption{Z3--Isabelle paired local-check outcomes for the same frozen
    Qwen2.5-7B-Instruct outputs. Rows denote Z3 and columns denote Isabelle,
    using \texttt{SAT}/\texttt{UNSAT}/\texttt{Other}. Each cell shows its
    count and percentage of that method's local check attempts.
    Isabelle requires a genuine Nitpick countermodel for \texttt{SAT}
    and an oracle-free kernel proof for \texttt{UNSAT}.
    \texttt{Other} retains input-processing failures and unresolved searches.}
    \label{fig:cross_solver_isabelle}
\end{figure}

Across 6,880 distinct queries, Isabelle constructs 2,574 kernel
proofs and 4,304 genuine countermodels. These confirm all 18,734
Z3-\texttt{UNSAT} steps and 12,198 of the 12,200 Z3-\texttt{SAT} steps,
with no contradictory decision among 30,932 jointly determinate pairs
(Figure~\ref{fig:cross_solver_isabelle}).
No Isabelle target-parsing, typing, or backend errors occur.

\paragraph{Proof verification rates.}
FVR and RVR are recomputed by replaying each backend's own accepted semantic
and rule-level prefixes. All semantic checks use the selected
backend, including auxiliary entailment and equivalence queries inside
\textsc{RuleCheck}.
The metrics retain the micro-averaged definitions in Appendix~\ref{app:evaluation_metrics}.
The four backends produce identical FVR and RVR numerators and denominators
for every response. Consequently, the rates also match within every
difficulty split. Table~\ref{tab:cross_solver_validation} reports the
Overall results. \method{} retains its verification gains over Backbone
and SFT under the alternative SMT implementations and native Isabelle proofs.

\begin{table}[!htbp]
    \setlength{\belowcaptionskip}{2.5pt}
    \caption{Cross-backend verification of the same frozen Qwen2.5-7B-Instruct
    outputs on ProverQA Overall. Values are percentages; every
    semantic check uses the indicated backend.}
    \label{tab:cross_solver_validation}
    \centering
    \fontsize{8}{8.5}\selectfont
    \setlength{\tabcolsep}{1pt}
    \renewcommand{\arraystretch}{1.2}
    \begin{tabular*}{\linewidth}{@{\extracolsep{\fill}}l*{8}{c}@{}}
        \toprule
        & \multicolumn{4}{c}{\textbf{FVR $\uparrow$}}
        & \multicolumn{4}{c}{\textbf{RVR $\uparrow$}} \\
        \cmidrule(lr){2-5}\cmidrule(lr){6-9}
        \textbf{Method} & Z3 & cvc5 & veriT & Isabelle & Z3 & cvc5 & veriT & Isabelle \\
        \midrule
        Backbone & 67.64 & 67.64 & 67.64 & 67.64
        & 18.91 & 18.91 & 18.91 & 18.91 \\
        SFT & 66.39 & 66.39 & 66.39 & 66.39
        & 19.94 & 19.94 & 19.94 & 19.94 \\
        \method{} & 70.17 & 70.17 & 70.17 & 70.17
        & 41.19 & 41.19 & 41.19 & 41.19 \\
        \bottomrule
    \end{tabular*}
\end{table}

\paragraph{Independent semantic review.}
We obtain blinded local-entailment judgments from GPT-5.5 on 2,000 steps:
667 Backbone, 667 SFT, and 666 \method{} steps, spanning 625 problems and
1,680 generated responses. 

GPT-5.5 receives only declared dependencies, their formulas, and the
conclusion. Generating-model identity and solver results are hidden.
It judges classical FOL entailment, returning \emph{valid}, \emph{invalid},
or \emph{uncertain} with a derivation, counterexample, or reason for
uncertainty. 

\begin{figure}[H]
    \centering
    \includegraphics[width=0.9\linewidth]{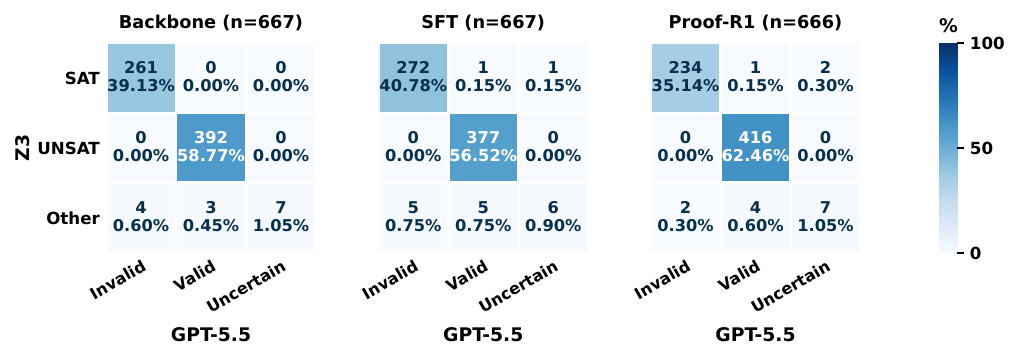}
    \caption{Z3--GPT-5.5 local semantic review of 2,000 sampled steps
    from frozen Qwen2.5-7B-Instruct outputs. Rows denote Z3 outcomes;
    columns denote GPT-5.5 judgments. \texttt{SAT} corresponds to
    \emph{invalid} and \texttt{UNSAT} to \emph{valid}.
    Each cell shows its count and percentage of reviewed steps. }
    \label{fig:review_agreement_gpt55}
\end{figure}

Figure~\ref{fig:review_agreement_gpt55} shows the corresponding
Z3--GPT-5.5 outcomes for each generating method.
The heatmap reports raw sample counts and within-method sample
percentages. Table~\ref{tab:independent_semantic_review} reports weighted
agreement for the 2,000-step sample.
GPT-5.5 endorses all 1,185 sampled \texttt{UNSAT} steps. Of 772
\texttt{SAT} steps, it labels 767 \emph{invalid}, three \emph{uncertain},
and two \emph{valid}. 

\begin{table}[!htbp]
    \setlength{\belowcaptionskip}{2pt}
    \caption{Agreement is measured on determinate Z3/cvc5 outcomes. The weighted percentage uses the sampling weights of the 2,000-step cohort.}
    \label{tab:independent_semantic_review}
    \centering
    \small
    \setlength{\tabcolsep}{5pt}
    \begin{tabular*}{\linewidth}{@{\extracolsep{\fill}}lrrrrrr@{}}
        \toprule
        \textbf{Reviewer} & \textbf{Number} & \textbf{Valid} & \textbf{Invalid} & \textbf{Uncertain} & \textbf{Agreed / reference} & \textbf{Weighted (\%)} \\
        \midrule
        GPT-5.5 & 2,000 & 1,199 & 778 & 23 & 1,952 / 1,957 & 99.73 \\
        \bottomrule
    \end{tabular*}
\end{table}

\paragraph{What do these independent checks establish?}
Together, the cross-backend verification and independent semantic review
establish that the verification gains of \method{} are not specific to Z3.
Replacing Z3 with cvc5, veriT, or Isabelle preserves every response-level
FVR and RVR, while Isabelle independently confirms
30,932 local judgments using native kernel proofs or genuine finite
countermodels. The blinded GPT-5.5 review further achieves 99.73\% weighted
agreement with the determinate Z3/cvc5 outcomes. Together, these results
demonstrate that \textit{the verification gains of \method{} are robust across
heterogeneous formal backends}.

\section{Rule Ontology}
\label{app:ontology}

\paragraph{RuleCheck.}
Each canonical rule $\rho\in\mathcal{R}$ is associated with a finite family
$\mathfrak{S}_{\rho}$ of admissible schemas.  A schema
$(\Gamma,\gamma)\in\mathfrak{S}_{\rho}$ consists of premise patterns
$\Gamma$ and a conclusion pattern $\gamma$.
Applying the resolver $\Phi_t^{\mathrm{rule}}(\cdot)$ from
Equation~\ref{eq:rule_verdict} to $D_t$ yields the dependency formulas passed
to \textsc{RuleCheck}, which we abbreviate as $\psi_t^{\mathrm{rule}}$:
\[
\psi_t^{\mathrm{rule}}
:=\Phi_t^{\mathrm{rule}}(D_t)
=\left\{
  \left(\mathcal{P}\cup\mathcal{S}_{t-1}^{\mathrm{rule}}\right)(d)
  \;\middle|\; d\in D_t
 \right\}.
\]
Using these resolved formulas, \textsc{RuleCheck} is defined by
\begin{equation}
\operatorname{RuleCheck}
  \left(\rho_t,\psi_t^{\mathrm{rule}},c_t\right)
=\ind\!\left[
  \rho_t\in\mathcal{R}
  \land
  \exists(\Gamma,\gamma)\in\mathfrak{S}_{\rho_t},\ \exists\sigma:
  \psi_t^{\mathrm{rule}}\models\sigma(\Gamma)
  \land c_t\equiv\sigma(\gamma)
\right].
\label{eq:rulecheck_definition}
\end{equation}
Here, $\sigma$ is a substitution over schema variables,
$\psi_t^{\mathrm{rule}}\models\sigma(\Gamma)$ means that every instantiated
premise pattern is established by the resolved dependency formulas, and
$\equiv$ denotes logical equivalence.  Consequently,
an unrecognized rule, an untrusted dependency, or the absence of a matching
schema yields a negative rule verdict.

\begin{table}[H]
  \setlength{\belowcaptionskip}{2pt}
    \caption{Predicate-logic rules covered by the ProverQA training data.}
    \label{tab:ontology}
    \centering
    \footnotesize
    \setlength{\tabcolsep}{3pt}
    \renewcommand{\arraystretch}{1.10}
    \resizebox{\linewidth}{!}{%
    \begin{tabular}{@{}p{0.53\linewidth}p{0.527\linewidth}@{}}
        \toprule
        \textbf{Rule} & \textbf{Symbolic form} \\
        \midrule
        \mbox{\textbf{IE}: \texttt{IMPLICATION\_ELIMINATION}}
        & $A\!\to\!B,\ A\vdash B$ \\
        \mbox{\textbf{MT}: \texttt{MODUS\_TOLLENS}}
        & $A\!\to\!B,\ \neg B\vdash\neg A$ \\
        \mbox{\textbf{XOI}: \texttt{EXCLUSIVE\_DISJUNCTION\_INTRODUCTION}}
        & $A,\ \neg B\vdash A\oplus B;\; \neg A,\ B\vdash A\oplus B$ \\
        \mbox{\textbf{XOE}: \texttt{EXCLUSIVE\_DISJUNCTION\_ELIMINATION}}
        & $A\oplus B,\ A\vdash\neg B;\; A\oplus B,\ \neg A\vdash B$ \\
        \mbox{\textbf{DS}: \texttt{DISJUNCTIVE\_SYLLOGISM}}
        & $A\lor B,\ \neg A\vdash B$ \\
        \mbox{\textbf{CI}: \texttt{CONJUNCTION\_INTRODUCTION}}
        & $A,\ B\vdash A\land B$ \\
        \mbox{\textbf{CE}: \texttt{CONJUNCTION\_ELIMINATION}}
        & $A\land B\vdash A;\; A\land B\vdash B$ \\
        \mbox{\textbf{UE}: \texttt{UNIVERSAL\_ELIMINATION}}
        & $\forall x\,\varphi(x)\vdash\varphi(a)$ \\
        \mbox{\textbf{GB}: \texttt{GOAL\_BINDING}}
        & $i_g=i(\widetilde y),\quad c_g=\widetilde y$ \\
        \bottomrule
    \end{tabular}%
    }
\end{table}

In the last row, $i(\widetilde y)$ denotes the terminal action identifier
assigned to the selected answer proposition.  \texttt{GOAL\_BINDING} is
accepted only when both that identifier and the conclusion match.

\section{Evaluation of \method{} on Out-of-Distribution Datasets}
\label{app:quantifier_coverage}

ProverQA is generated by a theorem prover, and its proof traces are constructed
from the predicate-logic ontology used for training.  FOLIO~\citep{han2024folio},
in contrast, is a human-authored natural-language reasoning benchmark whose
instances and annotations are independent of that generator.  We therefore
evaluate the frozen Qwen3-8B Backbone, SFT, and \method{} outputs on the FOLIO validation subset: 51 problems with three outputs per problem, or 153 outputs per model. 
In FOLIO, semantically valid derivations include universal and existential quantifier steps whose rule families are absent from the ProverQA training ontology. Let $\mathcal{R}_{\mathrm{unseen}}$ denote the seven unseen quantifier rule families listed in Table~\ref{tab:extended_rule_ontology}. The trusted inference system used for this FOLIO evaluation is therefore:
\begin{equation}
    \mathcal{R}_{\mathrm{FOLIO}}
    := \mathcal{R}_{\mathrm{train}}\cup\mathcal{R}_{\mathrm{unseen}}
    \supset \mathcal{R}_{\mathrm{train}}.
    \label{eq:folio_rule_system}
\end{equation}

To verify these deductions without changing the frozen outputs or the original
checker, a separate coverage checker operationalizes
$\mathcal{R}_{\mathrm{FOLIO}}$ by adding the seven guarded rule families in
Table~\ref{tab:extended_rule_ontology} to $\mathcal{R}_{\mathrm{train}}$.  Every formula, dependency, step
identifier, and \texttt{GOAL\_BINDING} action remains fixed.

\begin{table}[H]
    \setlength{\belowcaptionskip}{2pt}
    \caption{Training-time-unseen rule families and their representative symbolic forms.}
    \label{tab:extended_rule_ontology}
    \centering
    \footnotesize
    \setlength{\tabcolsep}{3pt}
    \renewcommand{\arraystretch}{1.10}
    \resizebox{\linewidth}{!}{%
    \begin{tabular}{@{}p{0.53\linewidth}p{0.527\linewidth}@{}}
        \toprule
        \textbf{Rule} & \textbf{Symbolic form} \\
        \midrule
        \mbox{\textbf{UIC}: \texttt{UNIVERSAL\_IMPLICATION\_CHAINING}}
        & $\forall x(A\!\to\!B),\ \forall x(B\!\to\!C)\vdash\forall x(A\!\to\!C)$ \\
        \mbox{\textbf{UPI}: \texttt{UNIVERSAL\_PROPOSITIONAL\_INFERENCE}}
        & $\forall x(A\!\to\!B),\ \forall x(A\!\to\!C)\vdash\forall x(A\!\to\!(B\land C))$ \\
        \mbox{\textbf{EI}: \texttt{EXISTENTIAL\_INTRODUCTION}}
        & $A(a)\vdash\exists x\,A(x);\; A(a),B(a)\vdash\exists x(A(x)\land B(x))$ \\
        \mbox{\textbf{ECE}: \texttt{EXISTENTIAL\_CONJUNCTION\_ELIMINATION}}
        & $\exists x(A\land B)\vdash\exists x\,A;\; \exists x(A\land G)\vdash G\ (x\notin\mathrm{FV}(G))$ \\
        \mbox{\textbf{EIE}: \texttt{EXISTENTIAL\_IMPLICATION\_ELIMINATION}}
        & $\exists xA,\ \forall x(A\!\to\!B)\vdash\exists xB$ \\
        \mbox{\textbf{ECI}: \texttt{EXISTENTIAL\_CONJUNCTION\_INTRODUCTION}}
        & $\exists xA,G\vdash\exists x(A\land G)\ (x\notin\mathrm{FV}(G))$ \\
        \mbox{\textbf{QN}: \texttt{QUANTIFIER\_NEGATION}}
        & $\neg\forall xA\Leftrightarrow\exists x\neg A;\; \neg\exists xA\Leftrightarrow\forall x\neg A$ \\
        \bottomrule
    \end{tabular}%
    }
\end{table}

Using the frozen outputs, we evaluate every step under
$\mathcal{R}_{\mathrm{FOLIO}}$. Let $\rho_t$ denote the rule family of the
current step. To isolate generalization to unseen rules, we group steps by
whether $\rho_t\in\mathcal{R}_{\mathrm{train}}$ or
$\rho_t\in\mathcal{R}_{\mathrm{unseen}}$. The group assignment depends only
on the current step.

\begin{table}[H]
    \setlength{\belowcaptionskip}{2pt}
    \caption{FVR/RVR (\%) under $\mathcal{R}_{\mathrm{FOLIO}}$ on
    frozen Qwen3-8B FOLIO outputs, grouped by whether the current rule belongs
    to $\mathcal{R}_{\mathrm{train}}$ or $\mathcal{R}_{\mathrm{unseen}}$.}
    \label{tab:folio_expanded_ontology}
    \centering
    \small
    \setlength{\tabcolsep}{8pt}
    \begin{tabular}{llrr}
        \toprule
        \textbf{Model} & \textbf{Rule group} & \textbf{FVR} & \textbf{RVR} \\
        \midrule
        Backbone & $\mathcal{R}_{\mathrm{train}}$ & 64.71 & 39.73 \\
                 & $\mathcal{R}_{\mathrm{unseen}}$ & 63.24 & 34.92 \\
        SFT      & $\mathcal{R}_{\mathrm{train}}$ & 62.02 & 39.77 \\
                 & $\mathcal{R}_{\mathrm{unseen}}$ & 67.50 & 45.71 \\
        \method{} & $\mathcal{R}_{\mathrm{train}}$ & \textbf{81.55} & \textbf{58.87} \\
                 & $\mathcal{R}_{\mathrm{unseen}}$ & \textbf{86.21} & \textbf{63.08} \\
        \bottomrule
    \end{tabular}
\end{table}

Under the expanded checker, \method{} achieves the highest FVR and RVR in
both groups: 81.55/58.87 for $\mathcal{R}_{\mathrm{train}}$ and 86.21/63.08
for $\mathcal{R}_{\mathrm{unseen}}$. On $\mathcal{R}_{\mathrm{unseen}}$, this improves over the
Backbone by 22.97 FVR points and 28.16 RVR points. Overall, \method{} reaches
59.70\% RVR under the expanded ontology. These results show that \emph{\method{} transfers from the
prover-generated training distribution to human-authored FOLIO and retains
verifiable reasoning under rule families absent from the training ontology.}

\section{Formal Specification and ProofCheck}
\label{app:specification}

\paragraph{Action specification.}
Each action proposes a proof-state transition whose admissibility is governed
by a formal specification:
\begin{align}
\operatorname{Spec}(a_t;\mathcal{S}_{t-1})={}&
\operatorname{WellFormed}(a_t)
\wedge\operatorname{TrustedDep}(D_t;\mathcal{S}_{t-1}) \nonumber\\
&\wedge\operatorname{SemanticValid}(D_t,c_t)
\wedge\operatorname{RuleFaithful}(D_t,c_t,\rho_t) \nonumber\\
&\wedge\operatorname{Admissible}(a_t).
\label{eq:specification}
\end{align}
\(\operatorname{WellFormed}\) checks the action schema, identifier, conclusion, and rule
field, while \(\operatorname{TrustedDep}\) restricts dependencies to original premises
or earlier conclusions admitted to the relevant verified state.
\(\operatorname{SemanticValid}\) requires entailment between the resolved dependency
formulas and the conclusion. \(\operatorname{RuleFaithful}\) requires the deduction to
instantiate the declared rule.
For ordinary steps, \(\operatorname{Admissible}\) is evaluated by
\(\operatorname{ProofCheck}\).
for the terminal action, it requires agreement between the identifier,
conclusion, and selected candidate.

\paragraph{ProofCheck.}
For ordinary actions, \textsc{ProofCheck} rejects repeated conclusions,
restatements of dependencies, and tautologies:
\begin{align}
v_t^{\mathrm{pro}}=\operatorname{ProofCheck}(a_t)={}&
\neg\operatorname{Repeated}(c_t)
\wedge\neg\operatorname{Restatement}(c_t,D_t)\nonumber\\
&\wedge\neg\operatorname{Tautological}(c_t).
\label{eq:progress_verdict}
\end{align}
Proof progress is separate from semantic soundness.
A repeated conclusion or tautology may be entailed without advancing the
proof, so nontriviality is not an entailment requirement.

\section{Structured Output Contract}
\label{app:schema}

\definecolor{schemakey}{RGB}{42,91,176}
\lstdefinestyle{structuredjson}{
  basicstyle=\ttfamily\small,
  columns=fullflexible,
  keepspaces=true,
  showstringspaces=false,
  frame=none,
  aboveskip=0pt,
  belowskip=0pt,
  lineskip=1pt,
  literate=
    {"id"}{{{\color{schemakey}\bfseries "id"}}}{4}
    {"dependencies"}{{{\color{schemakey}\bfseries "dependencies"}}}{14}
    {"conclusion"}{{{\color{schemakey}\bfseries "conclusion"}}}{12}
    {"rule"}{{{\color{schemakey}\bfseries "rule"}}}{6}
}

\newtcblisting{structuredjsonbox}{
  enhanced,
  listing only,
  listing engine=listings,
  listing options={style=structuredjson},
  width=\linewidth,
  colback=white,
  colframe=black,
  boxrule=0.6pt,
  arc=0pt,
  outer arc=0pt,
  left=3mm,
  right=3mm,
  top=2mm,
  bottom=2mm,
  before skip=0.5\baselineskip,
  after skip=0.6\baselineskip
}

\paragraph{Output organization.}
Following the rollout notation in Equation~\ref{eq:rollout}, the free-form
analysis region represents $z$, while the structured summary serializes
$(a_1,\ldots,a_T)$ as a JSON list. Its terminal action $a_g$ binds
$c_g=\tilde y$ for the selected candidate $\tilde y\in\mathcal{Y}$.

\paragraph{Action schema.}
As shown in Figure~\ref{fig:structured_output_example}, each action contains a unique identifier, a list of dependencies, one formal
conclusion, and its declared rule. For a candidate
set $\mathcal{Y}=\{y_1,\ldots,y_N\}$, the final action uses the identifier and
formal proposition of the selected option.
\begin{figure}[H]
\centering
\begin{structuredjsonbox}
[
  {"id": "s1", "dependencies": ["p1", "p2"],
   "conclusion": "Q(a)",
   "rule": "IMPLICATION_ELIMINATION"},
  {"id": "option_3", "dependencies": ["s1"],
   "conclusion": "Q(a)", 
   "rule": "GOAL_BINDING"}
]
\end{structuredjsonbox}
\caption{Example of the structured JSON proof-action sequence.}
\label{fig:structured_output_example}
\end{figure}

\section{Implementation Details}
\label{app:implementation}

We implement semantic verification with Z3 \citep{demoura2008z3}, and use
\textsc{RuleCheck} and \textsc{ProofCheck} for the corresponding rule
and proof-admissibility conditions.
Appendix~\ref{app:cross_solver_validation} reports paired cvc5, veriT, and Isabelle
checks of the frozen Qwen2.5-7B-Instruct evaluation outputs.
Policy optimization uses LoRA \citep{hu2022lora} within VERL/HybridFlow
\citep{sheng2024hybridflow}.

\subsection{Method-Specific Implementations}

To ensure a controlled comparison, all methods use the same data splits,
formal inputs, and structured output specification, and their final outputs
are scored by a common evaluator for answer accuracy and proof verifiability.
SFT, GRPO, PRoSFI, and \method{} use the same training configuration and LoRA
setup and are initialized from the same format-warm-up adapter. The methods
differ mainly in their inference procedures, training objectives, and
credit-assignment mechanisms.

\textbf{Backbone} is the pretrained model without task-specific training. Given
the common prompt, it directly generates a complete response containing a
reasoning trace, structured proof steps, and a final answer, without
candidate selection or verification feedback at inference time. The formal
verifier is used only for evaluation after generation and does not participate
in the generation process.

\textbf{LogicAgent} \citep{zhang2025logicagent} is a training-free neuro-symbolic
reasoning agent. It first uses an LLM to reason natural-language problems
from multiple semantic perspectives, delegates deduction to symbolic solvers,
and determines the final answer through reflection. We use the official
code repository.\footnote{\url{https://github.com/AI4SS/Logic-Agent}} To conform
to the common evaluation protocol, we only serialize its selected reasoning
process into the standard structured proof format, without using reference
answers or our verifier to alter its result.

\textbf{GoV} \citep{fang2026gov} is a training-free, verification-guided reasoning
agent. It first generates multiple candidate solutions and
organizes their reasoning steps as nodes in dependency graphs and then checks
the nodes in their declared dependency order and selects the final output from
the candidates that pass verification. Candidate selection uses GoV's own
LLM-based verifier. We use the official code repository.\footnote{\url{https://github.com/Frevor/Graph-of-Verification}}
To standardize the output, we only serialize the selected candidate into the
common structured proof format, without changing its selection result or
reasoning content.

\textbf{SFT} uses verified structured proof trajectories for supervised fine-tuning.
Given a problem and its formal representation, it learns through token-level
cross-entropy to generate a reasoning trace, canonicalized proof steps, and a
final answer. SFT performs neither online sampling nor additional feedback
based on the semantic or rule validity of generated proofs. Its supervision
therefore comes entirely from the provided demonstration trajectories.

\textbf{GRPO} \citep{shao2024deepseekmath} is initialized from the format-warm-up
adapter and samples a group of complete responses for each problem. Its reward
is determined jointly by output-format validity and final-answer correctness,
and relative rewards within each group are used to compute the advantage of
each response. This advantage is applied to the entire generated sequence and the
training reward does not separately check the semantic validity of intermediate
conclusions, the applicability of declared rules, or proof dependencies. We
implement GRPO using the official verl codebase.\footnote{\url{https://github.com/verl-project/verl}}

\textbf{PRoSFI} \citep{chen2026prosfi} is an RL method designed for natural-language
logical reasoning. It retains GRPO's grouped sampling and relative optimization,
is initialized from the format-warm-up adapter, and samples a group of complete
responses for each problem. For each response, it parses the structured proof,
checks intermediate conclusions with Z3 in dependency order, and assigns the
trajectory-level reward in Equation~\ref{eq:prosfi_reward}. Because the authors
have not released their code, we construct a faithful reimplementation following
the method described in the paper.

\begin{equation}
R_{\mathrm{PRoSFI}}=
\begin{cases}
1.0, & \text{correct answer; all steps verified},\\
0.3, & \text{correct answer; some steps unverified},\\
0.1, & \text{valid output format but incorrect answer},\\
0.0, & \text{invalid output format or other failure}.
\end{cases}
\label{eq:prosfi_reward}
\end{equation}

\subsection{Training Configuration}

The training configuration used in our experiments is summarized below.\footnote{Code is available at \url{https://anonymous.4open.science/r/Proof-R1/}.}

\begin{table}[H]
\centering
\caption{Training and rollout parameters. Unless stated otherwise, these
settings are shared by SFT, GRPO, PRoSFI, and \method{}.}
\label{tab:training_configuration}
\begingroup
\small
\setlength{\tabcolsep}{4pt}
\renewcommand{\arraystretch}{1.03}
\begin{tabularx}{\linewidth}{@{}p{0.24\linewidth}X>{\raggedright\arraybackslash}p{0.36\linewidth}@{}}
\toprule
\textbf{Category} & \textbf{Parameter} & \textbf{Value} \\
\midrule
LoRA & Rank & 8 \\
  & Alpha & 16 \\
  & Target modules & [Q, K, V, O, GATE, UP, DOWN] \\
\midrule
Rollout & Group size ($K$) & 8 \\
  & Sampling temperature & 0.8 \\
  & Top-$p$ & 0.95 \\
\midrule
Training & Training batch size & 8 \\
  & Sampled responses per batch & 64 \\
  & Total training steps & 1,098 \\
  & Training epochs & 3 \\
\midrule
Optimization & PPO epochs per update & 2 \\
  & PPO mini-batch size & 8 \\
  & Clip ratio ($\epsilon_-,\epsilon_+$) & $[0.20, 0.28]$ \\
  & Learning rate & $5\times10^{-6}$ \\
  & KL coefficient ($\lambda_{\mathrm{KL}}$) & 0.001 \\
  & Loss aggregation & seq-mean-token-sum \\
\midrule
\method{} Parameters & \multicolumn{2}{c}{\textit{Machine-Checkable Formal Verification (MCFV)}} \\
\cmidrule(l){2-3}
  & Initial value ($\operatorname{EMA}_\rho^{(0)}$) & 0.5 \\
  & EMA coefficient ($\beta$) & 0.9 \\
  & Clip range ($b_{\min},b_{\max}$) & $[0.30, 0.70]$ \\
\cmidrule(l){2-3}
  & \multicolumn{2}{c}{\textit{Verification-Aligned Optimization}} \\
\cmidrule(l){2-3}
  & Verification scale ($\lambda_v$) & 3 \\
  & Outcome scale ($\lambda_o$) & 1 \\
  & Goal-binding verification scale & 1 \\
  & No-positive-action weight & 0.1 \\
\midrule
Reasoning & Maximum prompt length & 4,096 \\
  & Maximum response length & 4,096 \\
  & Maximum model context length & 8,192 \\
\bottomrule
\end{tabularx}
\endgroup
\end{table}

\section{Additional Training Diagnostics}
\label{app:training_kl}

\begin{figure}[H]
    \centering
    \begin{minipage}[t]{0.49\linewidth}
        \centering
        \includegraphics[width=\linewidth]{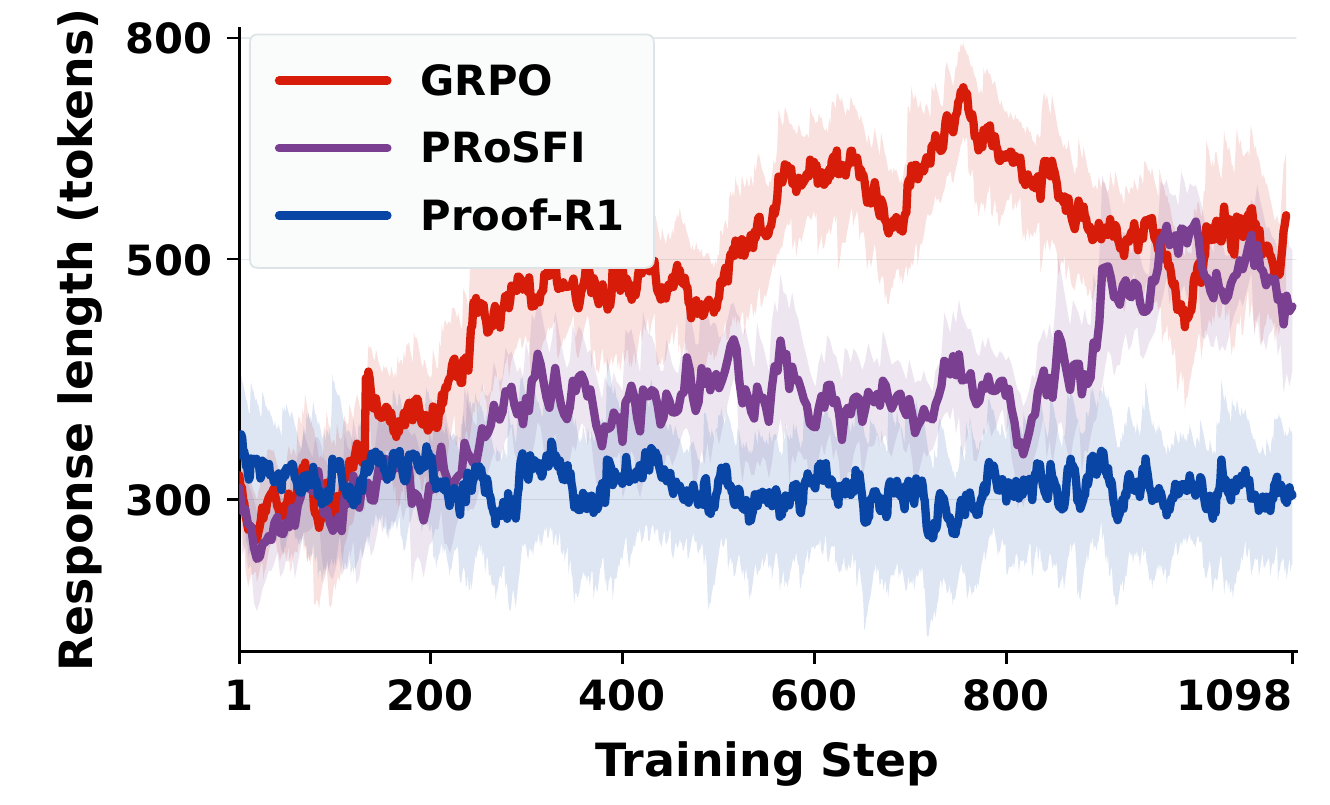}
    \end{minipage}\hfill
    \begin{minipage}[t]{0.49\linewidth}
        \centering
        \includegraphics[width=\linewidth]{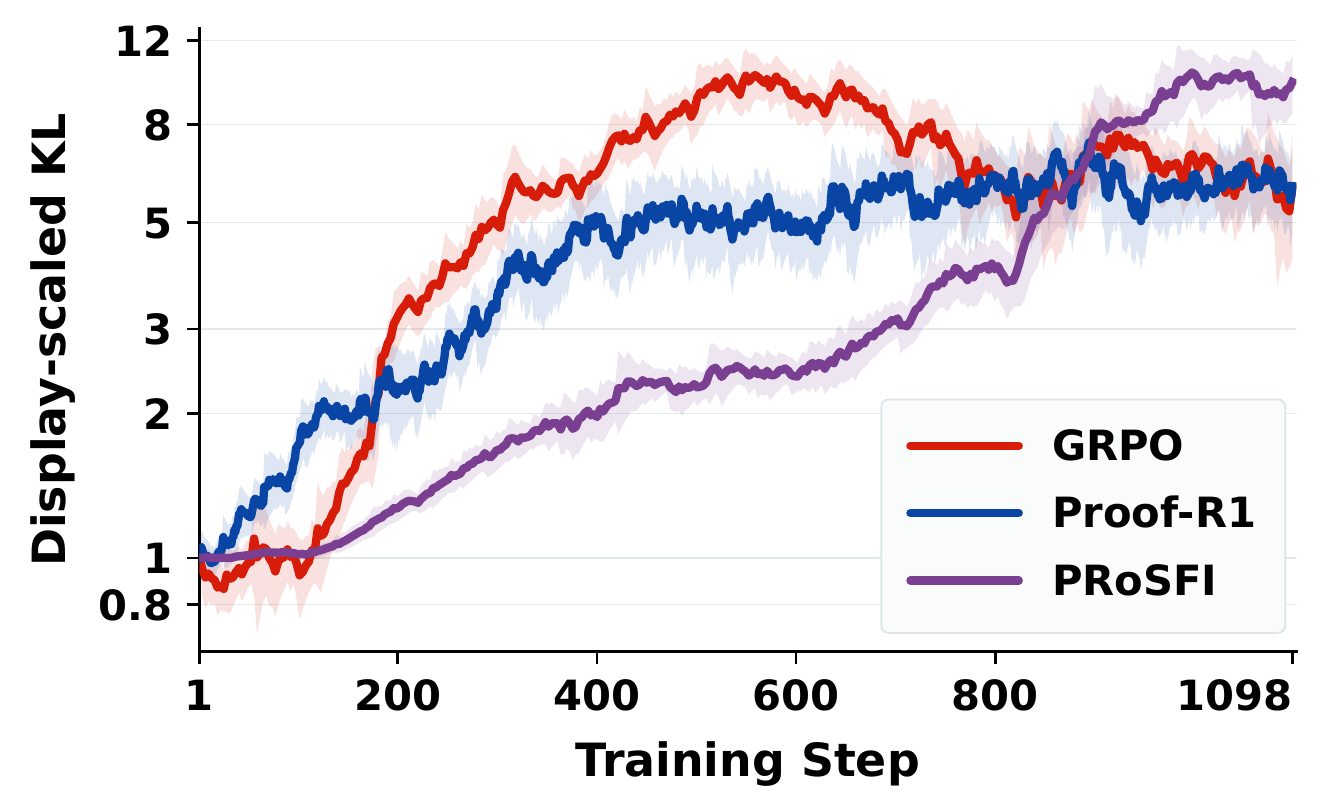}
    \end{minipage}
    \caption{Training trajectories of response length and policy divergence
    during reinforcement learning. Left: Mean response length of different
    methods on Qwen2.5-7B-Instruct during training. Right: Sequence-level KL
    divergence of the Qwen3-8B policy from the frozen reference policy during
    training. Solid lines denote EMA-smoothed trajectories, and shaded regions
    indicate one exponentially weighted local standard deviation.}
    \label{fig:additional_training_diagnostics}
\end{figure}

Figure~\ref{fig:additional_training_diagnostics} shows that, compared with the
response growth of GRPO and PRoSFI as training proceeds, \method{} maintains an
overall stable response length. This indicates that \emph{the training improvements
of \method{} do not rely on continually increasing response length, while the
policy gradually diverges from the initial reference policy under a stable
generation length}.

\section{Computational Cost}
\label{app:computational_cost}

All experiments use a PyTorch 2.5.1 base image with Python 3.12 on Ubuntu
22.04 and CUDA 12.4.  The machine provides four RTX 4090 GPUs with 48~GB of
memory each, 80 Intel(R) Xeon(R) Platinum 8470Q vCPUs, and 384~GB of system
memory.  Table~\ref{tab:computational_cost} reports wall-clock training time
per 100 optimization steps and the mean number of generated tokens consumed
per response at inference time. Training-free agents have zero training time.

\begin{table}[H]
    \setlength{\belowcaptionskip}{2pt}
    \caption{Computational cost on ProverQA.  Time is measured in hours per 100 optimization steps. Inference-time token consumption is the mean number of generated tokens per response.}
    \label{tab:computational_cost}
    \centering
    \small
    \setlength{\tabcolsep}{6pt}
    \renewcommand{\arraystretch}{1.12}
    \begin{tabular}{lrrrr}
        \toprule
        & \multicolumn{2}{c}{\textbf{Qwen2.5-7B-Instruct}}
        & \multicolumn{2}{c}{\textbf{Qwen3-8B}} \\
        \cmidrule(lr){2-3}\cmidrule(lr){4-5}
        \textbf{Method} & h/100 steps & tokens/response & h/100 steps & tokens/response \\
        \midrule
        Backbone   & 0.00 & 354.2   & 0.00 & 1,213.8 \\
        LogicAgent & 0.00 & 2,054.5 & 0.00 & 12,572.6 \\
        GoV        & 0.00 & 2,004.7 & 0.00 & 9,127.9 \\
        SFT        & 0.11 & 373.6   & 0.19 & 1,229.9 \\
        GRPO       & 2.64 & 836.4   & 4.70 & 1,176.3 \\
        PRoSFI     & 2.98 & 645.9   & 5.31 & 1,208.1 \\
        \textbf{\method{}} & 2.73 & 354.9 & 4.86 & 1,214.2 \\
        \bottomrule
    \end{tabular}
\end{table}

Across two backbones, \method{}, GRPO, and PRoSFI have comparable training speeds. \emph{\method{} consumes substantially fewer inference-time tokens than the LogicAgent and GoV baselines}.

\section{Prompt Template Used by \method{}}
\label{app:prompt}

The model receives a single-turn chat prompt consisting of a shared system
instruction and an instance-specific user message.  The structured-action
schema and user-message template are shared across format warm-up, policy
rollouts, and evaluation, while the inference-system block is instantiated for
the current dataset.  Qwen2.5-7B-Instruct and Llama-3.1-8B use the
explicit-thinking instruction below, whereas Qwen3-8B and GLM-Z1-9B-0414 use
their native thinking mode.

\definecolor{promptthink}{RGB}{72,143,38}
\definecolor{promptsummary}{RGB}{42,91,176}

\lstdefinestyle{methodprompt}{
  basicstyle=\rmfamily\small,
  columns=fullflexible,
  breaklines=true,
  breakatwhitespace=true,
  keepspaces=true,
  showstringspaces=false,
  frame=none,
  aboveskip=0pt,
  belowskip=0pt,
  lineskip=0pt,
  literate=
    {<think>}{{{\color{promptthink}\textless{}think\textgreater{}}}}{7}
    {</think>}{{{\color{promptthink}\textless{}/think\textgreater{}}}}{8}
    {<summary>}{{{\color{promptsummary}\textless{}summary\textgreater{}}}}{9}
    {</summary>}{{{\color{promptsummary}\textless{}/summary\textgreater{}}}}{10}
}

\newtcblisting{methodpromptbox}[1]{
  enhanced,
  breakable,
  listing only,
  listing engine=listings,
  listing options={style=methodprompt},
  width=\linewidth,
  colback=gray!6,
  colframe=black,
  colbacktitle=black,
  coltitle=white,
  fonttitle=\bfseries,
  title={#1},
  boxrule=0.8pt,
  arc=3mm,
  outer arc=3mm,
  left=3.5mm,
  right=3.5mm,
  top=2mm,
  bottom=2mm,
  toptitle=1.2mm,
  bottomtitle=1.2mm,
  before skip=0.5\baselineskip,
  after skip=0.7\baselineskip
}

The marker \texttt{[OUTPUT-MODE INSTRUCTION]} is replaced verbatim by one of
the two mode-specific blocks below.  The marker
\texttt{[INFERENCE SYSTEM BLOCK]} is instantiated with
$\mathcal{R}_{\mathrm{train}}$ for ProverQA and ProofWriter, and with
$\mathcal{R}_{\mathrm{FOLIO}}=\mathcal{R}_{\mathrm{train}}\cup
\mathcal{R}_{\mathrm{unseen}}$ for FOLIO.  The latter is formed by appending
the FOLIO extension block below to the training inference-system block.

\begin{methodpromptbox}{System Prompt}
You are a math reasoner. For each problem, you will receive a set of formal premises and a multiple-choice logical reasoning question.

[OUTPUT-MODE INSTRUCTION]

Each JSON object in <summary> must have exactly these fields:
- id: a unique identifier for this step.
- dependencies: a list of premise ids or earlier step ids.
- conclusion: exactly one formal formula.
- rule: the single inference rule applied, chosen from the supplied inference system.

[INFERENCE SYSTEM BLOCK]

The rule must describe the operation that produces the declared conclusion. If an implication first activates a compound consequent and the same action then selects one XOR/OR/AND branch, use the corresponding elimination rule, not IMPLICATION_ELIMINATION. Do not invent rule names, aliases, abbreviations, or alternative capitalization.

Use GOAL_BINDING for the final JSON object whose id is one of [option A id], [option B id], ..., or [option N id].

Critical requirements:
- Use "s1", "s2", "s3", etc. for intermediate steps.
- The last JSON object must be the final answer.
- The final answer id must exactly match one of the option ids shown in the problem.
- The final answer conclusion must match the formal statement of that option.
- The JSON array must be valid and parsable.
- Each step must use fewer than 5 dependencies.
- Do not output any additional text after </summary>.
\end{methodpromptbox}

\begin{methodpromptbox}{Explicit-Thinking Instruction (Qwen2.5-7B-Instruct and Llama-3.1-8B)}
Answer with exactly two parts:
1. Put brief natural-language reasoning inside <think>...</think>.
2. Put a JSON array of structured formal intermediate steps inside <summary>...</summary>.
\end{methodpromptbox}

\begin{methodpromptbox}{Native-Thinking Instruction (Qwen3-8B and GLM-Z1-9B-0414)}
Use the model's native thinking mode for brief natural-language reasoning.
After native thinking ends, output exactly one JSON array of structured formal intermediate steps inside <summary>...</summary>.
Do not add commentary between the native </think> and <summary> sections.
\end{methodpromptbox}

\needspace{8\baselineskip}
The training prompt instantiates the inference system
$\mathcal{R}_{\mathrm{train}}$ with the following canonical rules.
\begin{methodpromptbox}{Training Inference System Block ($\mathcal{R}_{\mathrm{train}}$)}
Canonical rule options:
Use exactly one of these canonical rule names:
- IMPLICATION_ELIMINATION: from A -> B and support for A, derive the entire consequent B.
- MODUS_TOLLENS: from A -> B and support for not B, derive not A (including a justified component of a compound A).
- EXCLUSIVE_DISJUNCTION_INTRODUCTION: from A and not B, or from not A and B, derive A XOR B.
- EXCLUSIVE_DISJUNCTION_ELIMINATION: from A XOR B plus one known branch or its negation, derive the forced other branch or its negation.
- DISJUNCTIVE_SYLLOGISM: from A OR B and not A derive B, or from A OR B and not B derive A.
- CONJUNCTION_INTRODUCTION: from A and B derive A AND B.
- CONJUNCTION_ELIMINATION: from A AND B derive A or derive B.
- UNIVERSAL_ELIMINATION: instantiate a universally quantified formula for one concrete entity.
- GOAL_BINDING: only for a final_answer action whose id and conclusion exactly match an answer option.
\end{methodpromptbox}

\needspace{8\baselineskip}
The FOLIO prompt extends $\mathcal{R}_{\mathrm{train}}$ to $\mathcal{R}_{\mathrm{FOLIO}}$ with $\mathcal{R}_{\mathrm{unseen}}$.
\begin{methodpromptbox}{Unseen Inference System Block ($\mathcal{R}_{\mathrm{unseen}}$)}
- UNIVERSAL_IMPLICATION_CHAINING: from forall x (A -> B) and forall x (B -> C), derive forall x (A -> C).
- UNIVERSAL_PROPOSITIONAL_INFERENCE: from forall x (A -> B) and forall x (A -> C), derive forall x (A -> (B AND C)).
- EXISTENTIAL_INTRODUCTION: from A(a), derive exists x A(x); premises sharing witness a may be combined under the existential.
- EXISTENTIAL_CONJUNCTION_ELIMINATION: from exists x (A AND B), derive exists x A; or derive a conjunct G that does not contain x free.
- EXISTENTIAL_IMPLICATION_ELIMINATION: from exists x A and forall x (A -> B), derive exists x B.
- EXISTENTIAL_CONJUNCTION_INTRODUCTION: from exists x A and G, derive exists x (A AND G) when x is not free in G.
- QUANTIFIER_NEGATION: apply the equivalences not forall x A iff exists x not A, and not exists x A iff forall x not A.
\end{methodpromptbox}

For each problem, the bracketed fields are populated from the corresponding
natural-language and formalized instance.  Premise formulas are converted to
the prefix notation consumed by the verifier.

\begin{methodpromptbox}{User Prompt Template}
Context:
1. [natural-language premise 1]. Formal statement: 'h1 : [formal premise 1]'.
2. [natural-language premise 2]. Formal statement: 'h2 : [formal premise 2]'.
...
N. [natural-language premise N]. Formal statement: 'hN : [formal premise N]'.

Question: [multiple-choice question]

Options:
A) [option A text]. Answer id: '[option A id]'. Formal statement: '[option A formula]'.
B) [option B text]. Answer id: '[option B id]'. Formal statement: '[option B formula]'.
...
N) [option N text]. Answer id: '[option N id]'. Formal statement: '[option N formula]'.

The correct option is:
\end{methodpromptbox}

\clearpage
\section{Case Study}
\label{app:case_study}

\newcommand{\casepremise}[1]{\texttt{p#1}}
\newlength{\casemodelwidth}
\newcommand{\casemodelheading}[2]{%
    \settowidth{\casemodelwidth}{\textbf{\method}\qquad}%
    \noindent\makebox[\casemodelwidth][l]{\textbf{#1}}#2\par}

\subsection{Case 1: A Negative Conclusion about Animal Care}
\label{app:case_alonzo}

\method{} connects a natural-language answer to an explicit sequence of checkable reasoning actions.
The case below presents a ProverQA example generated by \method{} with \mbox{Qwen3-8B}.

\begin{tcolorbox}[colback=black!2,colframe=black!2,
    boxrule=0pt,arc=0pt,left=8pt,right=8pt,top=8pt,bottom=8pt,
    before skip=9pt,after skip=10pt]
    \small
    \raggedright
    \definecolor{alonzoReuse}{RGB}{222,238,249}
    \definecolor{alonzoNegation}{RGB}{255,237,197}
    \newcommand{\alonzohighlight}[2]{{%
        \setlength{\fboxsep}{1.2pt}\colorbox{#1}{\strut #2}}}
    \begin{minipage}[t]{0.58\linewidth}
        \vspace{0pt}\raggedright
        \textbf{Premises $\mathcal{P}$}\par\vspace{4pt}
        \begin{description}[font=\normalfont\ttfamily,
            leftmargin=1.7em,labelwidth=1.3em,labelsep=0.4em,
            align=left,itemsep=3pt,parsep=0pt,topsep=0pt]
            \item[\casepremise{1}] Either Alonzo helps animals or harms animals,\newline
                \alonzohighlight{alonzoNegation}{but not both.}
            \item[\casepremise{2}] Alonzo has compassion.
            \item[\casepremise{3}] Alonzo loves wildlife.
            \item[\casepremise{4}] Anyone who loves wildlife and provides care\newline is helping animals.
            \item[\casepremise{5}] Anyone who feels empathy or has compassion\newline can provide care.
        \end{description}
    \end{minipage}\hfill
    \begin{minipage}[t]{0.38\linewidth}
        \vspace{0pt}\raggedright
        \textbf{Query $q$}\par\vspace{4pt}
        Alonzo does not harm animals.\par
        \vspace{9pt}
        \textbf{Candidates $\mathcal{Y}$}\qquad $\tilde y=y_1$\par\vspace{4pt}
        \begin{description}[font=\normalfont,
            leftmargin=1.7em,labelwidth=1.3em,labelsep=0.4em,
            align=left,itemsep=3pt,parsep=0pt,topsep=0pt]
            \item[$y_1$] Alonzo does not harm animals.
            \item[$y_2$] Alonzo harms animals.
            \item[$y_3$] It is uncertain whether Alonzo harms animals.
        \end{description}
    \end{minipage}\par

    \vspace{11pt}
    \casemodelheading{\method}{$\tilde y=y_1$}\vspace{5pt}
    \begingroup
    \fontsize{8}{10}\selectfont
    \newlength{\alonzolabelwidth}
    \settowidth{\alonzolabelwidth}{\textbf{\texttt{h\_goal\_true}}\quad}
    \newcommand{\alonzoreasoning}[2]{%
        \noindent\makebox[\alonzolabelwidth][l]{\textbf{\texttt{#1}}}%
        \parbox[t]{\dimexpr\linewidth-\alonzolabelwidth\relax}{%
            \raggedright #2\strut}\par\nobreak}
    \newcommand{\alonzoactionfields}[3]{%
        \noindent\makebox[0.20\linewidth][l]{$D_t=\texttt{#1}$}%
        \makebox[0.69\linewidth][l]{$c_t=#2$}%
        \makebox[0.11\linewidth][l]{$\rho_t=\mathrm{#3}$}\par}
    \alonzoreasoning{s1}{Alonzo has compassion, so \casepremise{5} implies that he provides care.}
    \alonzoactionfields{[\casepremise{2}, \casepremise{5}]}{\alonzohighlight{alonzoReuse}{\texttt{provides\_care(Alonzo)}}}{IE}

    \vspace{6pt}
    \alonzoreasoning{s2}{Combine his love of wildlife with the derived fact that he provides care.}
    \alonzoactionfields{[\alonzohighlight{alonzoReuse}{s1}, \casepremise{3}]}{\texttt{loves\_wildlife(Alonzo)}\land\texttt{provides\_care(Alonzo)}}{CI}

    \vspace{6pt}
    \alonzoreasoning{s3}{By \casepremise{4}, loving wildlife and providing care imply that he helps animals.}
    \alonzoactionfields{[s2, \casepremise{4}]}{\texttt{helps\_animals(Alonzo)}}{IE}

    \vspace{6pt}
    \alonzoreasoning{s4}{He helps animals; \casepremise{1} excludes helping and harming together, so he does not harm them.}
    \alonzoactionfields{[s3, \casepremise{1}]}{\alonzohighlight{alonzoNegation}{$\neg$\texttt{harms\_animals(Alonzo)}}}{XOE}

    \vspace{6pt}
    \alonzoreasoning{h\_goal\_true}{Bind this conclusion to the selected candidate $y_1$.}
    \alonzoactionfields{[s4]}{\neg\texttt{harms\_animals(Alonzo)}}{GB}
    \endgroup
\end{tcolorbox}

The blue highlights show how \method{} turns an intermediate result into a trusted dependency.
The conclusion \texttt{provides\_care(Alonzo)} at \texttt{s1} is admitted to the semantic and rule-verified states after its action passes verification (Section~\ref{sec:action_verification}).
The explicit \texttt{s1} reference in \texttt{s2} then reuses this fact to construct the conjunction required by \casepremise{4}, enabling \texttt{s3} to derive that Alonzo helps animals.
\textit{This link makes the source of the intermediate result and its role in subsequent reasoning directly checkable}.
The yellow highlights show how a constraint in the input supports the negative answer.
The phrase ``but not both'' in \casepremise{1} excludes simultaneous helping and harming.
Given the helping conclusion at \texttt{s3}, exclusive disjunction elimination at \texttt{s4} yields $\neg\texttt{harms\_animals(Alonzo)}$, which \texttt{h\_goal\_true} binds to candidate $y_1$.
The negative conclusion therefore has explicit supporting dependencies and a matching inference rule.
All five actions pass schema validation, UNSAT-based semantic verification, and \textsc{RuleCheck}.

\begingroup
\definecolor{caseReuse}{RGB}{222,238,249}
\definecolor{caseAnswer}{RGB}{255,237,197}
\definecolor{caseError}{RGB}{252,220,220}
\newcommand{\casehighlight}[2]{{%
    \setlength{\fboxsep}{1.2pt}\colorbox{#1}{\strut #2}}}
\newlength{\caselabelwidth}
\newcommand{\caseproofsetup}{%
    \fontsize{8}{10}\selectfont
    \setlength{\jot}{0pt}%
    \settowidth{\caselabelwidth}{\textbf{\texttt{h\_goal\_false}}\quad}}
\newcommand{\casereasoning}[2]{%
    \noindent\makebox[\caselabelwidth][l]{\textbf{\texttt{#1}}}%
    \parbox[t]{\dimexpr\linewidth-\caselabelwidth\relax}{%
        \raggedright #2\strut}\par\nobreak}
\newcommand{\caseactionfields}[3]{%
    \noindent\makebox[0.20\linewidth][l]{$D_t=\texttt{#1}$}%
    \makebox[0.69\linewidth][l]{$c_t=#2$}%
    \makebox[0.11\linewidth][l]{$\rho_t=#3$}\par}

\clearpage
\subsection{Case 2: Verified Dependencies for a Correct Answer}
\label{app:case_jayceon}

\MCFV{} makes the validity of an intermediate conclusion decisive for its reuse in a verified proof.
This ProverQA pair uses Qwen2.5-7B-Instruct on the same problem: the Backbone answers incorrectly, while \method{} derives the correct answer with verified actions.

\begin{tcolorbox}[colback=black!2,colframe=black!2,
    boxrule=0pt,arc=0pt,left=8pt,right=8pt,top=8pt,bottom=8pt,
    before skip=9pt,after skip=10pt]
    \small\raggedright
    \begin{minipage}[t]{0.58\linewidth}
        \vspace{0pt}\raggedright
        \textbf{Premises $\mathcal{P}$}\par\vspace{4pt}
        \begin{description}[font=\normalfont\ttfamily,
            leftmargin=1.7em,labelwidth=1.3em,labelsep=0.4em,
            align=left,itemsep=3pt,parsep=0pt,topsep=0pt]
            \item[\casepremise{1}] Jayceon does not have raw talent.
            \item[\casepremise{2}] If Jayceon is dedicated, then he can either achieve success or stay humble, but not both.
            \item[\casepremise{3}] Jayceon either has raw talent or is dedicated, but not both.
            \item[\casepremise{4}] Anyone who sets goals or works hard is dedicated.
            \item[\casepremise{5}] Jayceon achieves success.
            \item[\casepremise{6}] Superstar achieves success.
        \end{description}
    \end{minipage}\hfill
    \begin{minipage}[t]{0.38\linewidth}
        \vspace{0pt}\raggedright
        \textbf{Query $q$}\par\vspace{4pt}
        Jayceon stays humble.\par\vspace{9pt}
        \textbf{Candidates $\mathcal{Y}$}\qquad $y^*=y_2$\par\vspace{4pt}
        \begin{description}[font=\normalfont,
            leftmargin=1.7em,labelwidth=1.3em,labelsep=0.4em,
            align=left,itemsep=3pt,parsep=0pt,topsep=0pt]
            \item[$y_1$] Jayceon stays humble.
            \item[$y_2$] Jayceon does not stay humble.
            \item[$y_3$] It is uncertain whether Jayceon stays humble.
        \end{description}
    \end{minipage}\par

    \vspace{11pt}
    \casemodelheading{Backbone}{$\tilde y=y_1$}\vspace{5pt}
    \begingroup\caseproofsetup
    \casereasoning{s1}{Claims that Jayceon is not dedicated from the exclusive choice and his lack of talent.}
    \caseactionfields{[\casepremise{1}, \casepremise{3}]}{\casehighlight{caseError}{$\neg$\texttt{dedicated(Jayceon)}}}{\casehighlight{caseError}{XOI}}
    \vspace{6pt}
    \casereasoning{s2}{Uses this rejected conclusion with the success premise to claim that he stays humble.}
    \caseactionfields{[\casepremise{2},\casepremise{5},\casehighlight{caseError}{s1}]}{\casehighlight{caseError}{\texttt{stay\_humble(Jayceon)}}}{\mathrm{XOE}}
    \endgroup

    \vspace{11pt}
    \casemodelheading{\method}{$\tilde y=y_2$}\vspace{5pt}
    \begingroup\caseproofsetup
    \casereasoning{s1}{He has no raw talent, so the exclusive choice in \casepremise{3} establishes that he is dedicated.}
    \caseactionfields{[\casepremise{1}, \casepremise{3}]}{\casehighlight{caseReuse}{\texttt{dedicated(Jayceon)}}}{\mathrm{XOE}}
    \vspace{6pt}
    \casereasoning{s2}{Given his dedication and success, \casepremise{2} excludes staying humble.}
    \caseactionfields{[\casepremise{2},\casehighlight{caseReuse}{s1},\casepremise{5}]}{\casehighlight{caseAnswer}{$\neg$\texttt{stay\_humble(Jayceon)}}}{\mathrm{XOE}}
    \vspace{6pt}
    \casereasoning{h\_goal\_false}{Bind the negative conclusion to the correct candidate $y_2$.}
    \caseactionfields{[s2]}{\neg(\texttt{stay\_humble(Jayceon)})}{\mathrm{GB}}
    \endgroup
\end{tcolorbox}

The red highlights trace an unsupported conclusion into the Backbone's wrong answer.
Its first action passes schema validation, but the semantic check returns SAT and \textsc{RuleCheck} rejects the declared XOI rule: \casepremise{1} and \casepremise{3} entail \texttt{dedicated(Jayceon)}, not its negation.
The next action still cites \texttt{s1}, so verification reports a missing trusted dependency at the action that states the wrong answer.
The blue highlights show the corresponding verified transition in \method{}: exclusive disjunction elimination establishes dedication, which is admitted to both verified states and reused by \texttt{s2}.
This fact activates \casepremise{2}; together with success, it yields the yellow negative conclusion and the correct answer $y_2$.
All three actions pass schema, \emph{UNSAT-based semantic, and rule verification, yielding a continuous chain of trusted dependencies}.

\clearpage
\subsection{Case 3: Generating the Dependencies That Support the Answer}
\label{app:case_clay}

This ProverQA pair uses Qwen2.5-7B-Instruct on the same problem and sample.
The Backbone mixes two characters' facts; \method{} builds the dependencies establishing Clay's correct answer.

\begin{tcolorbox}[colback=black!2,colframe=black!2,
    boxrule=0pt,arc=0pt,left=8pt,right=8pt,top=6pt,bottom=6pt,
    before skip=9pt,after skip=10pt]
    \small\raggedright
    \begin{minipage}[t]{0.58\linewidth}
        \vspace{0pt}\raggedright
        \textbf{Premises $\mathcal{P}$}\par\vspace{4pt}
        \begin{description}[font=\normalfont\ttfamily,
            leftmargin=1.7em,labelwidth=1.3em,labelsep=0.4em,
            align=left,itemsep=2pt,parsep=0pt,topsep=0pt]
            \item[\casepremise{1}] Clay either learns magic theory or studies ancient tomes.
            \item[\casepremise{2}] Clay practices intricate spells.
            \item[\casepremise{3}] If Clay studies ancient tomes and practices intricate spells, then he can become a powerful mage.
            \item[\casepremise{4}] If Mitchell studies ancient tomes and practices intricate spells, then he can become a powerful mage.
            \item[\casepremise{5}] Clay does not learn magic theory.
            \item[\casepremise{6}] Mitchell practices intricate spells.
        \end{description}
    \end{minipage}\hfill
    \begin{minipage}[t]{0.38\linewidth}
        \vspace{0pt}\raggedright
        \textbf{Query $q$}\par\vspace{4pt}
        Clay can become a powerful mage.\par\vspace{9pt}
        \textbf{Candidates $\mathcal{Y}$}\qquad $y^*=y_1$\par\vspace{4pt}
        \begin{description}[font=\normalfont,
            leftmargin=1.7em,labelwidth=1.3em,labelsep=0.4em,
            align=left,itemsep=2pt,parsep=0pt,topsep=0pt]
            \item[$y_1$] Clay can become a powerful mage.
            \item[$y_2$] Clay cannot become a powerful mage.
            \item[$y_3$] It is uncertain whether Clay can become a powerful mage.
        \end{description}
    \end{minipage}\par

    \vspace{8pt}
    \casemodelheading{Backbone}{$\tilde y=y_2$}\vspace{4pt}
    \begingroup\caseproofsetup
    \casereasoning{s1}{Introduces a choice between Clay and Mitchell studying ancient tomes.}
    \caseactionfields{[\casepremise{1}, \casepremise{5}]}{\begin{aligned}[t]
        &(\texttt{study\_ancient\_tomes(Clay)}\\[-1pt]
        &\quad\lor\texttt{study\_ancient\_tomes(Mitchell)})
        \end{aligned}}{\casehighlight{caseError}{DS}}
    \vspace{3pt}
    \casereasoning{s2}{Combines the facts that Clay and Mitchell both practice intricate spells.}
    \caseactionfields{[\casepremise{2}, \casepremise{6}]}{\begin{aligned}[t]
        &(\texttt{practice\_intricate\_spells(Clay)}\\[-1pt]
        &\quad\land\texttt{practice\_intricate\_spells(Mitchell)})
        \end{aligned}}{\mathrm{CI}}
    \vspace{3pt}
    \casereasoning{s3}{Combines the two characters' conditions into a disjunction.}
    \caseactionfields{[\casehighlight{caseError}{s1}, s2]}{\begin{aligned}[t]
        &(\texttt{study\_ancient\_tomes(Clay)}\\[-1pt]
        &\quad\land\texttt{practice\_intricate\_spells(Clay)})\\[-1pt]
        &\lor(\texttt{study\_ancient\_tomes(Mitchell)}\\[-1pt]
        &\quad\land\texttt{practice\_intricate\_spells(Mitchell)})
        \end{aligned}}{\mathrm{XOI}}
    \vspace{3pt}
    \casereasoning{s4}{Cites both rules but introduces an unsupported disjunction with a negated Clay predicate.}
    \caseactionfields{[s3,\casepremise{3},\casepremise{4}]}{\begin{aligned}[t]
        &\casehighlight{caseError}{$\neg$(\texttt{become\_a\_powerful\_mage(Clay)})}\\[-1pt]
        &\lor(\texttt{become\_a\_powerful\_mage(Mitchell)})
        \end{aligned}}{\mathrm{IE}}
    \vspace{3pt}
    \casereasoning{s5}{Uses this disjunction and Mitchell's practice to claim that Clay cannot become a powerful mage.}
    \caseactionfields{[s4, \casepremise{6}]}{\casehighlight{caseError}{$\neg$(\texttt{become\_a\_powerful\_mage(Clay)})}}{\mathrm{MT}}
    \endgroup

    \vspace{8pt}
    \casemodelheading{\method}{$\tilde y=y_1$}\vspace{4pt}
    \begingroup\caseproofsetup
    \casereasoning{s1}{Clay does not learn magic theory, so \casepremise{1} implies that he studies ancient tomes.}
    \caseactionfields{[\casepremise{1}, \casepremise{5}]}{\casehighlight{caseReuse}{\texttt{study\_ancient\_tomes(Clay)}}}{\mathrm{DS}}
    \vspace{3pt}
    \casereasoning{s2}{Combine this fact with Clay's practice to establish both conditions required by \casepremise{3}.}
    \caseactionfields{[\casehighlight{caseReuse}{s1}, \casepremise{2}]}{\begin{aligned}[t]
        &\texttt{study\_ancient\_tomes(Clay)}\\[-1pt]
        &\land\texttt{practice\_intricate\_spells(Clay)}
        \end{aligned}}{\mathrm{CI}}
    \vspace{3pt}
    \casereasoning{s3}{Apply Clay's rule to these conditions, proving the correct candidate $y_1$.}
    \caseactionfields{[s2, \casehighlight{caseReuse}{\casepremise{3}}]}{\casehighlight{caseAnswer}{\texttt{become\_a\_powerful\_mage(Clay)}}}{\mathrm{IE}}
    \vspace{3pt}
    \casereasoning{h\_goal\_true}{Bind this conclusion to the selected candidate $y_1$.}
    \caseactionfields{[s3]}{\texttt{become\_a\_powerful\_mage(Clay)}}{\mathrm{GB}}
    \endgroup

    \vspace{8pt}
    \centering
    \begin{tikzpicture}[x=1.2cm,y=0.45cm,
        every node/.style={font=\fontsize{8}{10}\selectfont,inner sep=3pt},
        premise/.style={font=\fontsize{8}{10}\selectfont\ttfamily},
        action/.style={fill=caseReuse,font=\fontsize{8}{10}\selectfont\ttfamily}]
        \node[premise] (p1) at (0,0.6) {\casepremise{1}};
        \node[premise] (p5) at (0,-0.6) {\casepremise{5}};
        \node[action] (s1) at (1.2,0) {s1};
        \node[premise] (p2) at (2.4,0.9) {\casepremise{2}};
        \node[action] (s2) at (3,0) {s2};
        \node[premise] (p3) at (4.2,0.9) {\casepremise{3}};
        \node[action] (s3) at (4.8,0) {s3};
        \node[action,fill=caseAnswer] (goal) at (6.8,0) {h\_goal\_true};
        \node[premise,fill=black!20,text=black] (p4) at (2.4,-0.9) {\casepremise{4}};
        \node[premise,fill=black!20,text=black] (p6) at (4.2,-0.9) {\casepremise{6}};
        \draw[->] (p1)--(s1); \draw[->] (p5)--(s1);
        \draw[->] (s1)--(s2); \draw[->] (p2)--(s2);
        \draw[->] (s2)--(s3); \draw[->] (p3)--(s3);
        \draw[->] (s3)--(goal);
    \end{tikzpicture}\par
\end{tcolorbox}

Red marks the Backbone's rule mismatch at \texttt{s1}, the resulting untrusted dependency at \texttt{s3}, and the unsupported negation introduced at \texttt{s4} and propagated to \texttt{s5}.
Its first two conclusions are semantically valid, but the mixed path fails to prove the query.
The blue dependencies in \method{} connect Clay's facts to \casepremise{3}, yielding the yellow correct answer; all four actions pass schema, semantic, and rule verification.
Backward traversal in the candidate proof-certificate graph selects all four actions and premises \casepremise{1}, \casepremise{2}, \casepremise{3}, and \casepremise{5}; the gray-shaded \casepremise{4} and \casepremise{6} lie outside the closure.
\textit{This is the answer-supporting structure targeted by \ASDC{} training}.

\clearpage
\subsection{Case 4: Linking Verified Branches to the Answer}
\label{app:case_vance}

This ProverQA pair uses Qwen2.5-7B-Instruct on the same problem.
The Backbone leaves a verified fact disconnected from its answer, whereas \method{} joins two verified branches to establish the correct conclusion.

\begin{tcolorbox}[colback=black!2,colframe=black!2,
    boxrule=0pt,arc=0pt,left=8pt,right=8pt,top=6pt,bottom=6pt,
    before skip=9pt,after skip=10pt]
    \small\raggedright
    \begin{minipage}[t]{0.58\linewidth}
        \vspace{0pt}\raggedright
        \textbf{Premises $\mathcal{P}$}\par\vspace{4pt}
        \begin{description}[font=\normalfont\ttfamily,
            leftmargin=1.7em,labelwidth=1.3em,labelsep=0.4em,
            align=left,itemsep=2pt,parsep=0pt,topsep=0pt]
            \item[\casepremise{1}] Vance will make groundbreaking discoveries if he conducts rigorous research.
            \item[\casepremise{2}] If Ivan studies ocean depths, then he either makes groundbreaking discoveries or inspires new generations, but not both.
            \item[\casepremise{3}] Vance conducts rigorous research.
            \item[\casepremise{4}] Vance studies ocean depths.
            \item[\casepremise{5}] If Vance studies ocean depths, then he either makes groundbreaking discoveries or inspires new generations, but not both.
        \end{description}
    \end{minipage}\hfill
    \begin{minipage}[t]{0.38\linewidth}
        \vspace{0pt}\raggedright
        \textbf{Query $q$}\par\vspace{4pt}
        Vance inspires new generations.\par\vspace{9pt}
        \textbf{Candidates $\mathcal{Y}$}\qquad $y^*=y_2$\par\vspace{4pt}
        \begin{description}[font=\normalfont,
            leftmargin=1.7em,labelwidth=1.3em,labelsep=0.4em,
            align=left,itemsep=2pt,parsep=0pt,topsep=0pt]
            \item[$y_1$] Vance inspires new generations.
            \item[$y_2$] Vance does not inspire new generations.
            \item[$y_3$] It is uncertain whether Vance inspires new generations.
        \end{description}
    \end{minipage}\par

    \vspace{8pt}
    \casemodelheading{Backbone}{\hspace*{-1.2pt}\casehighlight{caseError}{$\tilde y=y_3$}}\vspace{4pt}
    \begingroup\caseproofsetup
    \settowidth{\caselabelwidth}{\textbf{\texttt{h\_goal\_uncertain}}\quad}
    \casereasoning{s1}{Derives the discovery fact from Vance's rigorous research.}
    \caseactionfields{[\casepremise{3}, \casepremise{1}]}{\casehighlight{black!20}{\texttt{make\_groundbreaking\_discoveries(Vance)}}}{\mathrm{IE}}
    \vspace{3pt}
    \casereasoning{s2}{Derives the exclusive choice from Vance's study of ocean depths.}
    \caseactionfields{[\casepremise{4}, \casepremise{5}]}{\begin{aligned}[t]
        &\texttt{make\_groundbreaking\_discoveries(Vance)}\\[-1pt]
        &\oplus\texttt{inspire\_new\_generations(Vance)}
        \end{aligned}}{\casehighlight{caseError}{XOI}}
    \vspace{3pt}
    \casereasoning{s3}{States that discovering and inspiring cannot both hold.}
    \caseactionfields{[\casehighlight{caseError}{s2}]}{\begin{aligned}[t]
        &\neg(\texttt{make\_groundbreaking\_discoveries(Vance)}\\[-1pt]
        &\quad\land\texttt{inspire\_new\_generations(Vance)})
        \end{aligned}}{\mathrm{MT}}
    \vspace{3pt}
    \casereasoning{h\_goal\_uncertain}{Selects uncertainty without using the discovery fact from \texttt{s1}.}
    \caseactionfields{[s2, s3]}{\casehighlight{caseError}{\texttt{inspire\_new\_generations(Vance)}}}{\mathrm{GB}}
    \endgroup

    \vspace{8pt}
    \casemodelheading{\method}{$\tilde y=y_2$}\vspace{4pt}
    \begingroup\caseproofsetup
    \settowidth{\caselabelwidth}{\textbf{\texttt{h\_goal\_uncertain}}\quad}
    \casereasoning{s1}{Vance's research establishes that he makes groundbreaking discoveries.}
    \caseactionfields{[\casepremise{3}, \casepremise{1}]}{\casehighlight{caseReuse}{\texttt{make\_groundbreaking\_discoveries(Vance)}}}{\mathrm{IE}}
    \vspace{3pt}
    \casereasoning{s2}{His study of ocean depths establishes the exclusive choice in \casepremise{5}.}
    \caseactionfields{[\casepremise{4}, \casepremise{5}]}{\begin{aligned}[t]
        &(\texttt{make\_groundbreaking\_discoveries(Vance)}\\[-1pt]
        &\quad\oplus\texttt{inspire\_new\_generations(Vance)})
        \end{aligned}}{\mathrm{IE}}
    \vspace{3pt}
    \casereasoning{s3}{Join both results: making discoveries excludes inspiring new generations.}
    \caseactionfields{[\casehighlight{caseReuse}{s1}, \casehighlight{caseReuse}{s2}]}{\casehighlight{caseAnswer}{$\neg$(\texttt{inspire\_new\_generations(Vance)})}}{\mathrm{XOE}}
    \vspace{3pt}
    \casereasoning{h\_goal\_false}{Bind this conclusion to the selected candidate $y_2$.}
    \caseactionfields{[s3]}{\neg(\texttt{inspire\_new\_generations(Vance)})}{\mathrm{GB}}
    \endgroup

    \vspace{8pt}
    \centering
    \begin{tikzpicture}[x=1.6cm,y=0.45cm,
        every node/.style={font=\fontsize{8}{10}\selectfont,inner sep=3pt},
        premise/.style={font=\fontsize{8}{10}\selectfont\ttfamily},
        action/.style={fill=caseReuse,font=\fontsize{8}{10}\selectfont\ttfamily}]
        \node[premise] (p3) at (0,1) {\casepremise{3}};
        \node[premise] (p1) at (0,0.2) {\casepremise{1}};
        \node[action] (s1) at (1.3,0.6) {s1};
        \node[premise] (p4) at (0,-0.6) {\casepremise{4}};
        \node[premise] (p5) at (0,-1.4) {\casepremise{5}};
        \node[action] (s2) at (1.3,-1) {s2};
        \node[action] (s3) at (3.8,-0.2) {s3};
        \node[action,fill=caseAnswer] (goal) at (5.5,-0.2) {h\_goal\_false};
        \node[premise,fill=black!20,text=black] (p2) at (3.8,-1.3) {\casepremise{2}};
        \draw[->] (p3)--(s1); \draw[->] (p1)--(s1);
        \draw[->] (p4)--(s2); \draw[->] (p5)--(s2);
        \draw[->] (s1)--(s3); \draw[->] (s2)--(s3);
        \draw[->] (s3)--(goal);
    \end{tikzpicture}\par
\end{tcolorbox}

The gray-highlighted Backbone conclusion at \texttt{s1} passes all four checks, yet it lies outside the dependency closure of \texttt{h\_goal\_uncertain}.
Its analysis likewise treats the exclusive choice as unresolved despite having established the discovery fact.
Red marks the rule mismatch at \texttt{s2}, the dependency at \texttt{s3} missing from the rule-verified state, and the final positive conclusion that fails semantic verification while the model selects $y_3$.
In \method{}, the blue references join \texttt{s1} and \texttt{s2} at \texttt{s3}: the established discovery resolves the exclusive choice and yields the yellow correct negative answer.
All four actions pass schema, semantic, and rule verification.
Backward traversal from the final node \texttt{s3} includes both branches and premises \casepremise{1}, \casepremise{3}, \casepremise{4}, and \casepremise{5}, leaving Ivan's \casepremise{2} outside the closure in gray.
\textit{\method{} can identify answer-relevant dependencies and integrate verified results across reasoning branches to construct a coherent proof of the correct answer}.
\endgroup

\stopcontents[appendices]

\end{document}